\documentclass[]{fairmeta}
\usepackage[table]{xcolor}
\usepackage{xspace}
\usepackage{booktabs,multirow,graphicx,adjustbox,amsmath,placeins,tabularx}
\usepackage{hyperref,url}
\usepackage{fvextra,needspace}
\definecolor{resultPurple}{HTML}{643E8C}
\newcommand{\keyresult}[1]{\textcolor{resultPurple}{\textbf{#1}}}
\newcommand{\dr}{\textbf{DecisionRank}\xspace}

\title{DecisionRank: \\On-Policy Distillation for Decision-Model Rerankers}

\author[1,2]{Yanran Wu}
\author[1]{Dian Yu}
\author[1]{Chongyang Bai}
\author[1]{Jieyi Zhang}
\author[1]{Rui Liu}
\author[1]{Tianyi Chen}
\author[1]{Shike Mei}
\author[1]{Hang Cui}
\author[1]{Frank Shyu}
\author[1]{Sandeep Pandey}
\author[1]{Luke Simon}
\author[2]{Chunwei Liu}
\author[1]{Xi Liu}

\affiliation[1]{Meta}
\affiliation[2]{Purdue University}

\abstract{

Decision models target structured tasks by directly predicting choices or scores, avoiding autoregressive decoding latency. However, transferring a generative LLM's ranking knowledge to a decision model remains challenging: the teacher expresses preferences through a ranked list of document ID sequences, while the student determines the ranking using scores from the first-token position. To bridge this gap, we introduce \dr, a decision model trained through item-space on-policy distillation (OPD). Item-space OPD constructs the student's rollout by sorting its document scores into a ranked sequence. A generative teacher takes the resulting ranking list as inputs and provides conditional feedback to directly supervise those scores. Combined with offline ranking supervision, \dr achieves state-of-the-art performance on both general and reasoning-intensive retrieval benchmarks such as TREC DL, BEIR, and BRIGHT. Across four benchmarks, item-space OPD consistently outperforms alternative training objectives. Even at 1.7B parameters, \dr surpasses its 27B generative teacher on TREC DL19--23. At 8B, it outperforms its generative listwise counterpart while running $24.5\times$ faster on TREC DL and $16.7\times$ faster on BRIGHT.

}

\date{\today}
\correspondence{\email{dianyu@meta.com}, \email{cbai@meta.com}, \email{chunwei@purdue.edu}}

\begin{document}

\maketitle

\begin{figure}[!htb]
\centering
\includegraphics[width=\linewidth]{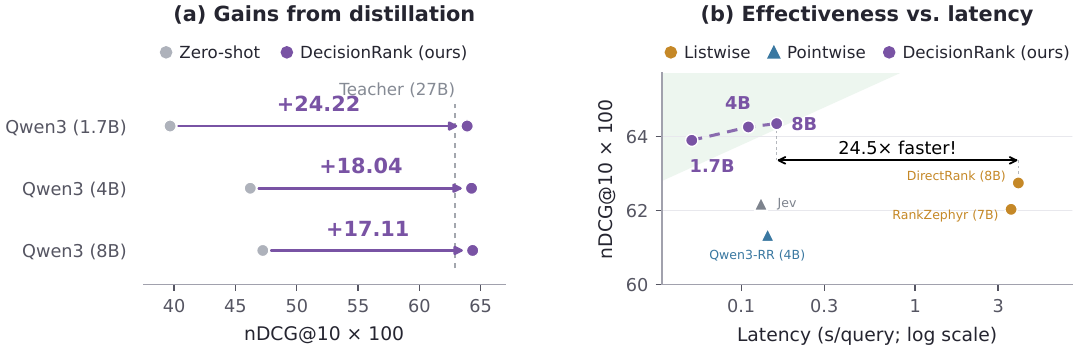}
\caption{\textbf{\dr: effective and efficient single-token reranking through distillation.} All methods rerank the same BM25 top-100 candidates on TREC DL19--23. (a) Distillation substantially improves ranking accuracy across all three student sizes, with each student outperforming the 27B generative teacher (62.91). Arrows indicate absolute gains in nDCG@10 × 100 over zero-shot scoring. (b) \dr students achieve favorable accuracy--latency tradeoffs compared with other rerankers. The gray triangle shows Jev, using median server-reported latency. Qwen3-RR denotes Qwen3-Reranker.}
\label{fig:homepage_teaser}
\end{figure}

\section{Introduction}

\begin{figure}[t]
\centering
\includegraphics[width=0.95\linewidth]{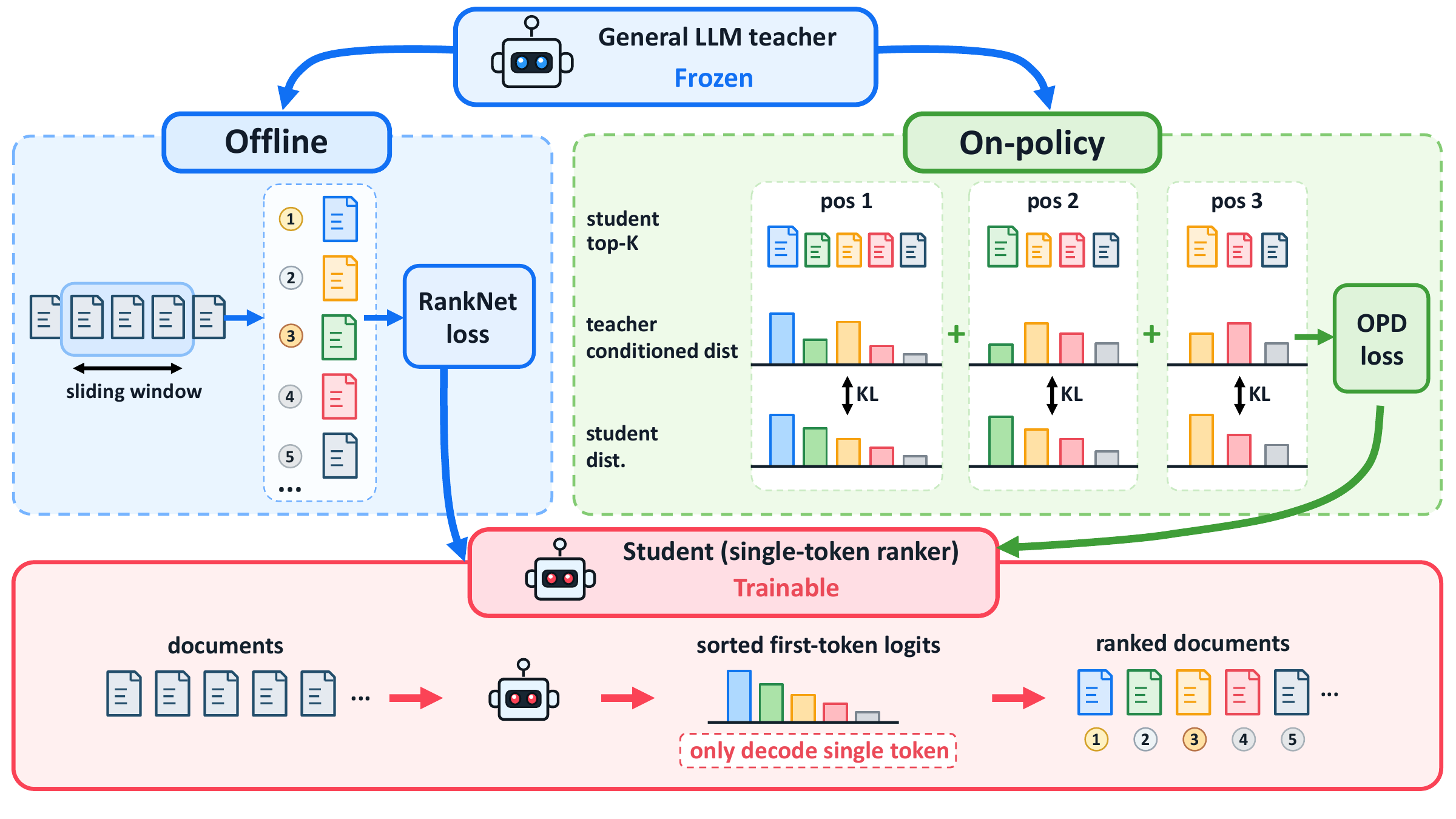}
\caption{Training \dr with offline rankings and on-policy feedback. A frozen generative teacher supplies offline rankings for RankNet (left) and conditional feedback along the student's current ranking (right). One teacher-forced forward pass provides the teacher's choice distributions, which are aligned with the student's scores renormalized over the remaining documents. Both objectives supervise the same first-token scores. The student's scoring procedure is the same during training and inference: a single forward pass produces the first-token scores used to rank all candidates.}
\label{fig:overview}
\end{figure}

Large language models (LLMs) have demonstrated strong performance in document reranking, particularly through listwise methods like RankGPT~\citep{sun2023chatgpt} that autoregressively generate candidate identifiers in descending order of relevance. However, this multi-step decoding incurs substantial inference latency. Decision models address this latency by directly predicting choices or scores without generating text, as illustrated by Jev \citep{almeida2026introducing} and AnyJev \citep{zhang2026anyjev}. In document reranking, single-token methods such as FIRST \citep{gangireddy2024first} follow the same principle: they sort first-token document-ID logits to obtain the ranking. Training these scores to capture the ranking preferences of capable generative teachers is therefore critical to effective single-token reranking.

Conventionally, ranking knowledge is transferred by training the student on static rankings generated offline by the teacher \citep{sun2023chatgpt,gangireddy2024first,schlatt2025rankdistillm} via learning-to-rank objectives. However, such fixed supervision cannot adapt to the student's evolving predictions, leaving the teacher unable to provide feedback on the student's current ranking decisions. While on-policy distillation (OPD) provides dynamic, state-aware feedback and has demonstrated notable success in autoregressive generation \citep{agarwal2024onpolicy,gu2024minillm}, effectively enabling it to complement offline ranking supervision in single-token rerankers remains challenging due to the structural mismatch between sequential feedback and single-pass scoring.

Specifically, standard token-level OPD relies on autoregressive rollouts to obtain conditional teacher feedback, whereas a single-token reranker represents the entire ranking through a static score vector in a single forward pass. Given this discrepancy, naive distillation strategies are insufficient: matching only the teacher's first-token distribution captures initial preferences but leaves subsequent conditional choices unused, while training the student to generate full ranking sequences creates a mismatch with the first-token scores used at inference. This raises our core question: \emph{How can we distill the sequential ranking preferences of an autoregressive teacher into a student that ranks all candidates in a single forward pass?}

To address this question, we propose \dr, a decision reranking model trained through item-space on-policy distillation (Figure~\ref{fig:overview}). Rather than relying on autoregressive student rollouts, we sort the student's first-token scores to construct ranking prefixes from its current top-$K$ candidates. A frozen generative teacher evaluates these prefixes in a single teacher-forced forward pass, providing conditional choice distributions over the remaining candidates at each ranking step. We obtain the corresponding student distributions by normalizing its first-token scores over the same remaining candidates, and align them with the teacher distributions using reverse Kullback--Leibler (KL) divergence in item space. This directly supervises the student's first-token scores used for inference, while allowing teacher feedback to adapt as the student's ranking evolves. Because the alignment operates over candidate identities rather than vocabulary tokens, the teacher and student need not share a vocabulary or model architecture. We jointly optimize this on-policy objective with offline RankNet supervision derived from teacher-generated rankings, combining global ordering preferences with conditional feedback on the student's current ranking. At inference, the student preserves its single-pass efficiency, ranking the full candidate set without autoregressive generation.

We conduct extensive experiments across four benchmarks covering both general and reasoning-intensive retrieval tasks. Our method achieves a strong effectiveness--latency tradeoff compared with existing LLM-based rerankers. As shown in Figure~\ref{fig:homepage_teaser}, all three distilled students (1.7B, 4B, and 8B) outperform the 27B generative teacher on TREC DL19--23 \citep{craswell2020trec2019,craswell2025trec2023}. Notably, our 8B student also surpasses the listwise DirectRank-8B while running $24.5\times$ faster on TREC DL under the same input length settings. Our method consistently improves ranking effectiveness over offline RankNet supervision across all four benchmarks and outperforms alternative training objectives, including Offline KL, autoregressive SFT, autoregressive OPD, and Listwise GRPO \citep{shao2024deepseekmath,prabhakar2026onpolicy}. Ablation studies further demonstrate the complementary benefits of offline RankNet supervision and on-policy distillation, with their combination achieving the strongest performance. Beyond ranking effectiveness, our 8B students yield lower pairwise calibration error than their Qwen3.8 and Gemma teachers on TREC DL and BRIGHT, without a calibration-specific training objective (Section~\ref{sec:rq1_calibration}).

The contributions of this paper are threefold:
\begin{itemize}
    \item We introduce item-space on-policy distillation and combine it with offline ranking supervision to distill the ranking ability of generative LLM teachers into \dr, a decision model for document reranking.
    \item We show that \dr achieves competitive performance across general and reasoning-intensive retrieval benchmarks, with substantially lower inference latency than generative listwise rerankers.
    \item We demonstrate generalizability across teacher families and student sizes, with consistent improvements over zero-shot students across all 4 benchmarks.
\end{itemize}

\section{Method}
\label{sec:method}
\dr is a decision model for document reranking that learns from a generative teacher. It combines offline ranking supervision with on-policy feedback (Figure~\ref{fig:overview}). In this section, we describe the 3 key components of our method. \textbf{First-Token Ranking and Item-Space Alignment} uses first-token document-ID logits as ranking scores and aligns teacher and student choice distributions over the same candidate items. \textbf{On-Policy Conditional Feedback} uses the student's current ranking as a sequence of document selections. The teacher evaluates the remaining documents given those already selected, providing feedback to supervise the student's scores. \textbf{Offline Supervision and Joint Training} combines this feedback with global ordering preferences from offline teacher rankings.

\subsection{First-Token Ranking and Item-Space Alignment}
Given a query $q$ and retrieved candidates $D=(d_1,\ldots,d_N)$ with $N\leq100$, the student reads the query and all candidates together. We assign each candidate $d_i$ a distinct document ID $v_i$ from a pool of 100 single-token IDs in the student vocabulary. Let $s_i$ denote the logit of $v_i$ at the first output position. Sorting these scores yields the student's ranking:
\begin{equation}
    \pi = (\pi_1, \pi_2, \ldots, \pi_N) = \operatorname{argsort}_{i\in\{1,\ldots,N\}}^{\downarrow} s_i.
    \label{eq:first_token_ranking}
\end{equation}
Each candidate document $d_i$ is an \emph{item} in our ranking problem. The teacher uses numerical document IDs, while the student uses single-token IDs. Although these IDs can have different token representations, they refer to the same candidate items. We align the two models' predictions for each item and compare their choice probabilities over the same candidate set. This item-space alignment enables distillation across model families.

\subsection{On-Policy Conditional Feedback}
To obtain on-policy feedback, we use the student's current ranking to construct a sequence of selection decisions and query the teacher at the resulting states. We select the top-$K$ candidates $S=\{\pi_1,\ldots,\pi_K\}$ and follow their order in $\pi$. At step $t$, let $h_t=(\pi_1,\ldots,\pi_{t-1})$ record the student's earlier selections and $R_t=S\setminus\{\pi_1,\ldots,\pi_{t-1}\}$ contain the remaining candidates. The student's scores define a Plackett--Luce choice distribution over $R_t$:
\begin{equation}
    p_{\theta,t}(i)=\frac{\exp(s_i)}{\sum_{j\in R_t}\exp(s_j)},\qquad i\in R_t.
    \label{eq:student_choice}
\end{equation}
Every step reuses the same first-token scores; only the remaining set changes. The trajectory itself is the deterministic ranking induced by those scores.

We construct the teacher prompt from $q$ and the selected documents $D_S$, retaining their first-stage retrieval order. We express the student's ranking using the teacher's numerical IDs and use it as the teacher-forced answer sequence. At step $t$, the teacher evaluates the remaining items given the earlier student selections $h_t$. Let $a_{T,t}(i)>0$ denote its conditional score for selecting item $d_i$. We map these scores to the corresponding candidate items and normalize over the remaining set $R_t$:
\begin{equation}
    p_{T,t}(i)=\frac{a_{T,t}(i)}{\sum_{j\in R_t}a_{T,t}(j)},\qquad i\in R_t.
    \label{eq:teacher_choice}
\end{equation}
Both distributions are therefore defined over the same remaining items at every step.
All conditional distributions are extracted in one teacher-forced forward pass over the student-induced answer sequence. This supplies feedback on the ranking states visited by the current student without an autoregressive rollout.

We align the student's choices with the teacher's conditional choices using reverse KL:
\begin{equation}
    \ell_t = D_{\mathrm{KL}}\!\left(p_{\theta,t}\,\Vert\,p_{T,t}\right)
    = \sum_{i\in R_t}p_{\theta,t}(i)
      \log\frac{p_{\theta,t}(i)}{p_{T,t}(i)}.
    \label{eq:conditional_distillation}
\end{equation}

The on-policy objective $\mathcal{L}_{\mathrm{on}}$ averages these step divergences over a designated set of steps $T_{\mathrm{step}}$:
\begin{equation}
    \mathcal{L}_{\mathrm{on}} = \frac{1}{|T_{\mathrm{step}}|}\sum_{t\in T_{\mathrm{step}}} \ell_t.
    \label{eq:on_policy_loss}
\end{equation}
In joint training, we use $T_{\mathrm{step}}=\{2,\ldots,K-1\}$. Starting from step 2 focuses the feedback on choices conditioned on earlier student selections, while offline supervision provides global ordering guidance. The final step is omitted because only one candidate remains and its KL divergence is zero.

Within each training step, the selected candidates and their order are fixed when computing the loss. Gradients update the student through its first-token scores, encouraging its choice distributions to agree with the teacher's conditional preferences. As training updates these scores, we recompute the ranking and obtain teacher feedback on the student's current selections.

\subsection{Offline Supervision and Joint Training}
The frozen teacher also provides full rankings generated offline using a sliding window with window size $W$ and stride $s$. We supervise the student's first-token scores on these static teacher rankings using RankNet \citep{burges2005learning}:
\begin{equation}
    \mathcal{L}_{\mathrm{RN}} =
    \frac{\sum_{(i,j)\in \mathcal{P}} w_{ij} \log\left(1 + \exp(s_j - s_i)\right)}
    {\sum_{(i,j)\in \mathcal{P}} w_{ij}},
    \label{eq:ranknet_loss}
\end{equation}
where $\mathcal{P}=\{(i,j) \mid d_i \succ d_j\}$ contains pairs where $d_i$ is ranked above $d_j$ by the teacher. Let $r_i^T$ denote the teacher rank of $d_i$. We use the rank utility $u(r)=1/\log_2(1+r)$ and weight each pair by $w_{ij}=u(r_i^T)-u(r_j^T)$.

The complete training objective combines global ordering supervision with conditional feedback on the student's current selections:
\begin{equation}
    \mathcal{L}=\mathcal{L}_{\mathrm{RN}}+\lambda\mathcal{L}_{\mathrm{on}},
    \label{eq:joint_objective}
\end{equation}
where $\lambda > 0$ controls the contribution of on-policy feedback. Offline RankNet supervises the global ordering across the full candidate set. On-policy distillation adds conditional feedback on the student's current top-$K$ selections. Both objectives optimize the same first-token scores used at inference, while the teacher remains frozen throughout training.

\paragraph{Single-pass inference.}
During inference, the trained student processes the query $q$ and the full candidate set of $N$ retrieved documents in a single forward pass, extracts their first-token identifier logits, and returns the ranking permutation $\pi$ via Equation~\ref{eq:first_token_ranking}. The full candidate ordering is obtained by sorting scores, with no autoregressive sequence generation.

\section{Experiments}
We organize our experiments around four research questions:
\begin{enumerate}
\item[\textbf{RQ1.}] How does our method compare with existing rerankers?
\item[\textbf{RQ2.}] Which training objective works best for ranking with first-token logits?
\item[\textbf{RQ3.}] Does combining offline and on-policy supervision improve ranking?
\item[\textbf{RQ4.}] Does our method generalize across teachers and student sizes?
\end{enumerate}

\subsection{Experimental Setup}
\subsubsection{Training}
\label{sec:training_setup}
\paragraph{Training data.}
We train on 10,000 MS MARCO queries from Rank-DistiLLM \citep{schlatt2025rankdistillm} and 6,000 synthetic queries from ReasonEmbed \citep{chen2025reasonembed}. For ReasonEmbed, we sample 500 queries from each of the 12 categories in the released dataset. For Rank-DistiLLM, we rerank the candidate documents with the corresponding teacher to construct offline ranking supervision.

\paragraph{Students and teachers.}
Our students are initialized from Qwen3 at 1.7B, 4B, and 8B parameters \citep{yang2025qwen3}. Qwen3.8-27B \citep{qwen2026qwen38} serves as the teacher for the main experiments. All teachers remain frozen and provide both offline rankings and on-policy feedback. To evaluate generalizability across teachers, we also distill from Gemma-4-31B \citep{gemmateam2026gemma4} and ReasonRank-32B \citep{liu2026reasonrank} (Section~\ref{sec:rq4}). Training and inference details are provided in Appendix~\ref{app:training_details}.

\begin{table}[!t]
\centering
\small
\setlength{\tabcolsep}{4pt}
\renewcommand{\arraystretch}{1.10}
\caption{\textbf{Comparison with existing rerankers.} Our Qwen3 students are distilled from Qwen3.8-27B. Scores are nDCG@10 $\times 100$, averaged across datasets within each benchmark. Bold and underlined values mark the best and second-best results, respectively. $^\dagger$The gray rows report Jev through its proprietary API and are excluded from this highlighting.}
\label{tab:rq1_summary}
\begin{adjustbox}{max width=\linewidth}

\definecolor{rqOneOurs}{HTML}{EEE8F7}
\definecolor{rqOneQwenText}{HTML}{0052CC}
\definecolor{rqOneApiText}{HTML}{666666}
\begin{tabularx}{\linewidth}{lc*{5}{>{\centering\arraybackslash}X}}
\toprule
& & \multicolumn{2}{c}{General Retrieval} & \multicolumn{2}{c}{Reasoning-Intensive Retrieval} \\
\cmidrule(lr){3-4}\cmidrule(lr){5-6}
\multicolumn{1}{c}{Method} & Backbone & TREC DL & BEIR & BRIGHT & R2MED & Avg. \\
\midrule
First-stage retrieval & -- & 39.26 & 43.74 & 30.56 & 24.69 & 34.56\\
\midrule
\multicolumn{6}{l}{\textit{Pointwise}} \\
Qwen3-Reranker-4B & \textcolor{rqOneQwenText}{Qwen3} & 61.34 & 55.70 & 26.64 & 34.70 & 44.60\\
Qwen3-Reranker-8B & \textcolor{rqOneQwenText}{Qwen3} & 60.42 & 55.55 & 27.21 & 36.98 & 45.04\\
TFRank-4B & \textcolor{rqOneQwenText}{Qwen3} & 57.82 & 53.86 & 31.05 & 37.36 & 45.02 \\
TFRank-8B & \textcolor{rqOneQwenText}{Qwen3} & 57.96 & 53.94 & 31.92 & 36.13 & 44.99\\
ERank-4B & \textcolor{rqOneQwenText}{Qwen3} & 59.51 & 54.32 & 33.97 & 40.23 & 47.01\\
ERank-14B & \textcolor{rqOneQwenText}{Qwen3} & 58.65 & 54.81 & 33.62 & 40.76 & 46.96\\
ERank-32B & \textcolor{rqOneQwenText}{Qwen3} & 59.06 & 54.78 & 35.33 & 41.20 & 47.59\\
DirectRank-8B & \textcolor{rqOneQwenText}{Qwen3} & 58.70 & 52.52 & 33.37 & 40.43 & 46.26 \\
\midrule
\multicolumn{6}{l}{\textit{Listwise}} \\
RankZephyr-7B & Zephyr-$\beta$ & 62.07 & 52.90 & 26.70 & 25.00 & 41.67\\
FIRST-7B & Zephyr-$\beta$ & 61.65 & 45.67 & 22.30 & 24.58 & 38.55\\
FirstMistral-7B & Mistral-v0.3 & 61.19 & 54.89 & 26.11 & 28.38 & 42.64\\
RankMistral-7B & Mistral-v0.3 & 61.60 & 54.82 & 29.63 & 32.53 & 44.65\\
Rank-R1-7B & Qwen2.5 & 59.93 & 53.80 & 28.63 & 33.97 & 44.08\\
Rank-R1-14B & Qwen2.5 & 60.34 & 55.50 & 31.96 & 39.97 & 46.94\\
DirectRank-8B & \textcolor{rqOneQwenText}{Qwen3} & 62.92 & 55.78 & \underline{35.87} & \textbf{44.50} & 49.77\\
GroupRank-7B & Qwen2.5 & 58.77 & 52.16 & 31.81 & 35.36 & 44.53 \\
GroupRank-32B & Qwen2.5 & 59.31 & 54.37 & 35.07 & 40.15 & 47.23\\
\midrule
\textcolor{rqOneApiText}{Jev-score (pointwise)$^{\dagger}$} & \textcolor{rqOneApiText}{API} & \textcolor{rqOneApiText}{62.18} & \textcolor{rqOneApiText}{55.48} & \textcolor{rqOneApiText}{32.79} & \textcolor{rqOneApiText}{44.75} & \textcolor{rqOneApiText}{48.80} \\
\textcolor{rqOneApiText}{Jev-choice (listwise)$^{\dagger}$} & \textcolor{rqOneApiText}{API} & \textcolor{rqOneApiText}{57.96} & \textcolor{rqOneApiText}{52.71} & \textcolor{rqOneApiText}{36.29} & \textcolor{rqOneApiText}{44.12} & \textcolor{rqOneApiText}{47.77}\\
\midrule
\rowcolor{rqOneOurs}
\dr{}-1.7B & \textcolor{rqOneQwenText}{Qwen3} & 63.89 & 56.65 & 34.08 & 38.45 & 48.27 \\
\rowcolor{rqOneOurs}
\dr{}-4B & \textcolor{rqOneQwenText}{Qwen3} & \underline{64.25} & \underline{57.60} & 35.61 & 42.39 & \underline{49.96}\\
\rowcolor{rqOneOurs}
\dr{}-8B & \textcolor{rqOneQwenText}{Qwen3} & \textbf{64.34} & \textbf{57.98} & \textbf{36.86} & \underline{43.55} & \textbf{50.68}\\
\bottomrule
\end{tabularx}

\end{adjustbox}
\end{table}

\subsubsection{Evaluation}
\label{sec:evaluation_setup}
\paragraph{Benchmarks and candidates.}
We evaluate reranking effectiveness on 4 benchmarks spanning general and reasoning-intensive retrieval.
\textbf{\textcircled{\scriptsize 1} General retrieval benchmarks.}
We evaluate on the 5 TREC DL years from DL19 to DL23 \citep{craswell2020trec2019,craswell2025trec2023}, using the MS MARCO passage collections \citep{bajaj2018msmarco}: DL19--20 use the v1 collection and DL21--23 use v2. To assess out-of-domain general retrieval, we evaluate on 7 BEIR datasets \citep{thakur2021beir}: TREC-COVID, NFCorpus, DBPedia, SciFact, Signal-1M, TREC-News, and Robust04. Both TREC DL and BEIR use BM25 top-100 candidates.
\textbf{\textcircled{\scriptsize 2} Reasoning-intensive retrieval benchmarks.}
BRIGHT comprises 12 tasks spanning diverse domains \citep{su2025bright}. We use top-100 candidates retrieved by ReasonIR-8B \citep{shao2025reasonir} with the released GPT-4 reasoning queries, while rerankers receive the original query. R2MED evaluates out-of-domain transfer to medical reasoning retrieval across 8 datasets \citep{zhang2026r2med}, using top-100 candidates retrieved by E5-Mistral-7B-Instruct \citep{wang2024improving}. Out-of-domain evaluation here refers to transfer beyond our distillation training data.

\paragraph{Evaluation metric.}
We report normalized discounted cumulative gain at rank 10 (nDCG@10), which measures relevance among the top 10 results while assigning greater weight to higher ranks:
\begin{equation}
\operatorname{nDCG@10}(q,\pi)
=\frac{1}{\operatorname{IDCG@10}(q)}
\sum_{r=1}^{10}\frac{g_q(d_{\pi_r})}{\log_2(r+1)}.
\label{eq:eval_ndcg}
\end{equation}
Here, $\pi=(\pi_1,\ldots,\pi_N)$ orders the $N$ candidate documents for query $q$: $\pi_r$ is the index of the document placed at rank $r$, and $d_{\pi_r}$ is that document. The gain $g_q(d)$ quantifies the relevance of document $d$ to query $q$ using the benchmark's annotated relevance judgments (qrels). Irrelevant documents receive zero gain, and higher relevance grades receive larger gains, following the benchmark's gain mapping. $\operatorname{IDCG@10}(q)$ is the same discounted sum for the ideal ranking, computed from the benchmark's full relevance judgments for $q$. We first average over queries within each dataset, then take an unweighted mean over the 5 TREC DL years, 7 BEIR datasets, 12 BRIGHT tasks, or 8 R2MED datasets. All scores are multiplied by 100; higher is better.

\subsection{RQ1: How does our method compare with existing rerankers?}
\label{sec:rq1}
We compare our method with pointwise and listwise rerankers. Pointwise methods score each query--document pair independently, while listwise methods consider multiple documents together. We evaluate ranking effectiveness across all 4 benchmarks, then examine inference efficiency and score calibration on TREC DL and BRIGHT.

\subsubsection{Ranking Effectiveness}
\label{sec:rq1_effectiveness}
Table~\ref{tab:rq1_summary} compares our three student sizes with existing rerankers on general retrieval and reasoning-intensive retrieval. We also include Jev~\citep{almeida2026introducing} (\texttt{jev-1.14.0}) through its pointwise Score and listwise Choice interfaces. Score ranks documents by expected relevance grade, while Choice ranks all 100 candidates by their returned choice probabilities.

\paragraph{General retrieval.}
\keyresult{All three student sizes outperform the compared baselines on both TREC DL and BEIR.} Our 8B student improves over the strongest baseline, listwise DirectRank, by 1.42 points on TREC DL and 2.20 points on BEIR. The advantage already appears at 1.7B: our smallest student exceeds every compared baseline on both benchmark averages, including models with up to 32B parameters.

\paragraph{Reasoning-intensive retrieval.}
\keyresult{Our 8B student achieves the highest average score on BRIGHT, and both 4B and 8B students outperform all open-weight pointwise baselines on BRIGHT and R2MED.} The 8B student reaches 36.86 on BRIGHT, improving over listwise DirectRank by 0.99 points. On R2MED, it achieves 43.55, compared with 44.50 for listwise DirectRank. With just 16,000 training queries, our method distills the teacher's ability to rerank reasoning-intensive queries and transfers this ability to out-of-domain medical retrieval.

\subsubsection{Inference Efficiency}
\label{sec:rq1_efficiency}
\begin{figure}[!htbp]
\centering
\includegraphics[width=\linewidth]{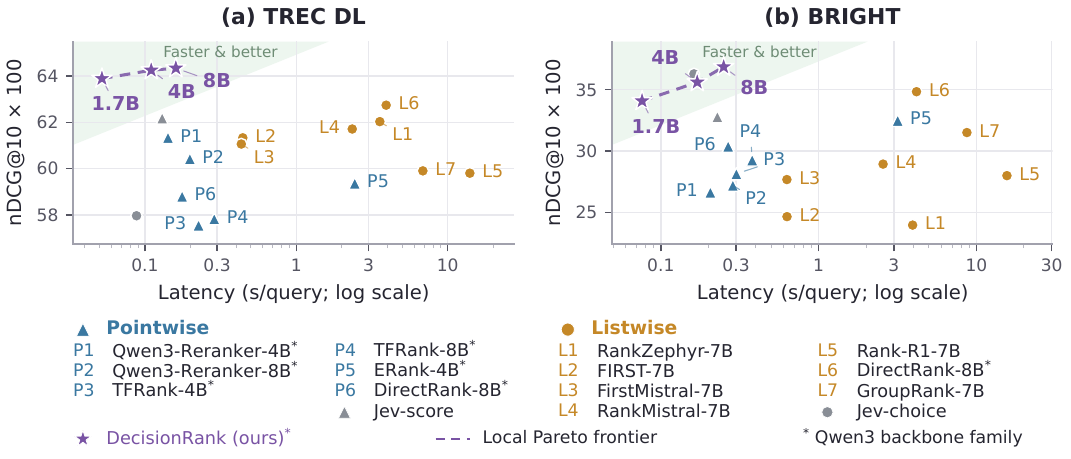}
\caption{\textbf{Ranking effectiveness and inference latency on TREC DL and BRIGHT.} Local models rerank 100 candidates using their original inference configurations, with passage limits of 128 tokens on TREC DL and 384 on BRIGHT. Scores are benchmark-average nDCG@10 $\times 100$. Local latency is measured on an NVIDIA B200; gray markers show Jev with median server-reported latency. The latency axis is logarithmic. Dashed lines show the Pareto frontier among locally evaluated models. Superscript $^{*}$ marks models from the Qwen3 backbone family.}
\label{fig:rq1_pareto}
\end{figure}

Figure~\ref{fig:rq1_pareto} compares ranking effectiveness against single-query latency on TREC DL and BRIGHT to assess the efficiency of full-list reranking. Latency measurement details are provided in Appendix~\ref{app:training_details}.

\keyresult{Our three students form the effectiveness--latency Pareto frontier among locally evaluated models on both benchmarks.} Even our 1.7B student achieves higher nDCG@10 and lower latency than Jev in both its pointwise and listwise modes on TREC DL. It also achieves the lowest latency on both benchmarks. On BRIGHT, our 4B student offers comparable ranking effectiveness and latency to Jev's listwise mode. Our 8B student outperforms listwise DirectRank on both benchmarks while running $24.5\times$ faster on TREC DL and $16.7\times$ faster on BRIGHT.

\subsubsection{Score Calibration}
\label{sec:rq1_calibration}
A ranker's confidence is more useful if it reflects how often its preferences are correct. We compare \dr with its generative teachers and the proprietary decision model Jev. We also include DirectRank-8B and ERank-32B, representing the two strongest open-weight pointwise reranker families by average ranking effectiveness in Table~\ref{tab:rq1_summary}. We evaluate pairwise score calibration~\citep{peng2023stabilizing} using expected calibration error (ECE)~\citep{guo2017calibration}. For pairs of documents with different relevance grades, each model assigns a probability to its predicted preference. We group these predictions into bins and measure the gap between average confidence and empirical accuracy:
\begin{equation}
\operatorname{ECE}
=\sum_{b=1}^{B}\frac{|B_b|}{M}
\left|\operatorname{acc}(B_b)-\operatorname{conf}(B_b)\right|,
\qquad M=\sum_{b=1}^{B}|B_b|.
\label{eq:calibration_ece}
\end{equation}
Here, $B_b$ contains the document pairs in bin $b$, $\operatorname{acc}(B_b)$ is the fraction of correct pairwise predictions, and $\operatorname{conf}(B_b)$ is their mean predicted confidence. Lower ECE indicates closer agreement between confidence and accuracy.

\begin{figure}[!htbp]
\centering
\includegraphics[width=0.78\linewidth]{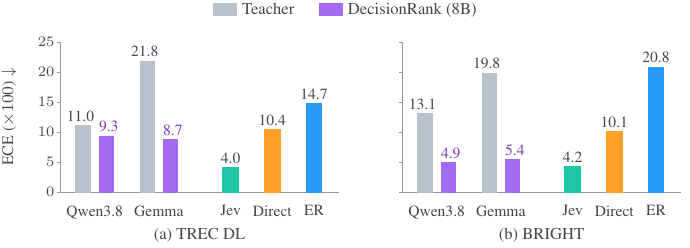}
\caption{\textbf{Pairwise score calibration on (a) TREC DL and (b) BRIGHT.} Each panel compares Qwen3.8-27B and Gemma-4-31B with their respective Qwen3-8B \dr students, followed by Jev, pointwise DirectRank-8B (Direct), and ERank-32B (ER). We report ECE $\times100$ over all evaluation queries; lower is better.}
\label{fig:rq1_calibration}
\end{figure}

As shown in Figure~\ref{fig:rq1_calibration}, \keyresult{both \dr students achieve lower ECE than their teachers on TREC DL and BRIGHT}, measured over document pairs across all evaluation queries. On BRIGHT, their ECE \textbf{is comparable to that of the proprietary decision model Jev} and lower than that of pointwise DirectRank and ERank. \dr achieves this level of calibration through ranking supervision alone, without a dedicated calibration objective. Details of the probability calculations and ECE aggregation, along with additional results for 4B students, are provided in Appendix~\ref{app:calibration}.

\FloatBarrier
\subsection{RQ2: Which training objective works best for ranking with first-token logits?}
\label{sec:rq2}
Here, we ask which form of teacher supervision best improves the first-token reranking scores used at inference. We compare five additional objectives on top of a common RankNet loss: offline logit distillation (Offline KL), autoregressive supervised fine-tuning (AR SFT), autoregressive on-policy distillation (AR OPD), Listwise GRPO, and our item-space on-policy distillation (Item-space OPD).

\paragraph{RankNet.}
RankNet is a classic pairwise learning-to-rank loss \citep{burges2005learning}. Following FIRST \citep{gangireddy2024first}, we use a weighted RankNet objective (Equation~\ref{eq:ranknet_loss}) as the foundation for distillation. The following objectives each add a different form of supervision to this common foundation.

\paragraph{+ Offline KL.}
Offline KL matches the student's first-token distribution to a fixed teacher distribution. Before training, we run the teacher on the full candidate set and cache its first-position distribution over candidate IDs. We then minimize forward KL, $D_{\mathrm{KL}}(p_T\Vert p_S)$, with both distributions normalized over all candidate IDs at temperature 1.

\paragraph{+ Autoregressive SFT (AR SFT).}
AR SFT teaches the student to generate the teacher's ranking as an autoregressive sequence. We apply teacher-forced cross-entropy to the ranking tokens, following the joint ranking and language-model supervision in FIRST \citep{gangireddy2024first}. When prompted to rank 100 candidates in a single response, LLMs produce incomplete or invalid rankings for most queries. We therefore only generate the first 20 candidate IDs from the teacher's offline ranking and use them as the SFT target. The student learns to generate this ranking sequence while RankNet loss supervises the full candidate set on the first token logits.

\paragraph{+ Autoregressive OPD (AR OPD).}
AR OPD applies distillation at each candidate-ID position in a student-generated ranking, following the GKD paradigm \citep{agarwal2024onpolicy}. The student first generates 20 distinct candidate rankings autoregressively. We provide the teacher with the query and the selected candidates, and append the student-generated ranking as a teacher-forced response. At each ID position, both models predict the next candidate conditioned on the same student-generated prefix. We minimize reverse KL, $D_{\mathrm{KL}}(p_S\Vert p_T)$, between the student's and teacher's next-ID distributions at steps 1--19, with both distributions renormalized over the remaining IDs within the sampled set of 20.

\paragraph{+ Listwise GRPO.}
Listwise GRPO optimizes student-sampled rankings with an nDCG@10 reward, aligning with the final evaluation metric. The teacher's offline ranking provides the reference for computing this reward. We adapt Group Relative Policy Optimization (GRPO) \citep{shao2024deepseekmath} to ranking, drawing on the ranking-policy formulation of \citet{prabhakar2026onpolicy}. The student's first-token scores define a stochastic ranking policy: a probability distribution over candidate orderings. For a query $q$ and candidates $D=(d_1,\ldots,d_N)$, let $\pi=(\pi_1,\ldots,\pi_{|\pi|})$ denote a sampled ordering of distinct candidate indices, where $\pi_t$ is the index selected at position $t$ and $|\pi|$ is the ranking length. The policy assigns probability $P_\theta(\pi\mid q,D)$ to this ordering, where $\theta$ denotes the student parameters. Using its first-token scores $s_i$, we define this probability through the Plackett--Luce model:
\begin{equation}
P_\theta(\pi\mid q,D)=\prod_{t=1}^{|\pi|}
\frac{\exp(s_{\pi_t})}{\sum_{j\in\{1,\ldots,N\}\setminus\{\pi_1,\ldots,\pi_{t-1}\}}\exp(s_j)}.
\label{eq:grpo_policy}
\end{equation}
Each factor is the probability of selecting candidate $\pi_t$ from those not yet selected; their product gives the probability of the sampled ordering. All selections reuse the same first-token scores from one student forward pass.

From this policy, we draw $G=8$ rankings $\pi^{(1)},\ldots,\pi^{(G)}$ per query using Gumbel sampling. If candidate $i$ has rank $\rho_i$ in the teacher's offline ordering of $N$ candidates, we assign it relevance $y_i=(N-\rho_i)/(N-1)$, with gain $2^{y_i}-1$. We compute an nDCG@10 reward for each sampled ranking and standardize the rewards within the group:
\begin{equation}
R_g=\operatorname{nDCG@10}(\pi^{(g)};y),
\qquad A_g=\frac{R_g-\bar R}{\sigma_R},
\label{eq:grpo_advantage}
\end{equation}
where $\bar R$ and $\sigma_R$ are the mean and standard deviation of the eight rewards. We update the student using these advantages, favoring rankings with above-average rewards. Rewards are computed from the same offline teacher ranking used by RankNet.

\paragraph{+ Item-space OPD (ours).}
We design Item-space OPD to obtain corrective teacher feedback on the student's ranking decisions. At each training step, the student's first-token logits select and order its top-$K$ candidates. We provide these candidates to the teacher and use the student's ranking as teacher-forced answer prefixes. A single forward pass yields the teacher's conditional next-item distributions along these prefixes. We compute reverse KL in item space: teacher and student distributions are aligned by candidate IDs (Equation~\ref{eq:conditional_distillation}). This feedback guides the student's ordering toward the teacher's preferences. As the student's ranking evolves, the teacher is queried on the updated candidates and prefixes. This type of loss directly applies to the first-token logits and is well-aligned with the student's inference procedure.

\begin{table}[t]
\centering
\small
\setlength{\tabcolsep}{4pt}
\renewcommand{\arraystretch}{1.15}
\caption{\textbf{Comparison of different training objectives.} All configurations use a Qwen3-8B student and a Qwen3.8-27B teacher, retain the RankNet loss, and use first-token inference. We report nDCG@10 $\times 100$, averaged equally across all datasets within each benchmark; parentheses show changes from RankNet.}
\label{tab:rq2_objectives}
\begin{adjustbox}{max width=\linewidth}

\begingroup
\definecolor{rqTwoGain}{HTML}{28C77A}
\definecolor{rqTwoLoss}{HTML}{D58A50}
\setlength{\fboxsep}{1.2pt}
\begin{tabular}{@{}l*{4}{r@{\hspace{2pt}}l}@{}}
\toprule
& \multicolumn{4}{c}{General Retrieval} & \multicolumn{4}{c}{Reasoning-Intensive Retrieval} \\
\cmidrule(lr){2-5}\cmidrule(l){6-9}
Training objective & \multicolumn{2}{c}{TREC DL} & \multicolumn{2}{c}{BEIR} & \multicolumn{2}{c}{R2MED} & \multicolumn{2}{c}{BRIGHT} \\
\midrule
RankNet & 63.44 &  & 57.39 &  & 43.06 &  & 36.06 &  \\
\quad + Offline KL & 61.99 & \colorbox{rqTwoLoss!45.00!white}{{\scriptsize $(-1.45)$}} & 57.15 & \colorbox{rqTwoLoss!45.00!white}{{\scriptsize $(-0.24)$}} & 42.80 & \colorbox{rqTwoLoss!37.15!white}{{\scriptsize $(-0.26)$}} & 36.85 & \colorbox{rqTwoGain!44.54!white}{{\scriptsize $(+0.79)$}} \\
\quad + AR SFT & 63.15 & \colorbox{rqTwoLoss!15.40!white}{{\scriptsize $(-0.29)$}} & 57.31 & \colorbox{rqTwoLoss!20.33!white}{{\scriptsize $(-0.08)$}} & 43.45 & \colorbox{rqTwoGain!37.45!white}{{\scriptsize $(+0.39)$}} & 36.40 & \colorbox{rqTwoGain!23.73!white}{{\scriptsize $(+0.34)$}} \\
\quad + AR OPD & 63.30 & \colorbox{rqTwoLoss!11.57!white}{{\scriptsize $(-0.14)$}} & 57.27 & \colorbox{rqTwoLoss!26.50!white}{{\scriptsize $(-0.12)$}} & 43.41 & \colorbox{rqTwoGain!34.43!white}{{\scriptsize $(+0.35)$}} & 36.42 & \colorbox{rqTwoGain!24.65!white}{{\scriptsize $(+0.36)$}} \\
\quad + Listwise GRPO & 63.75 & \colorbox{rqTwoGain!20.74!white}{{\scriptsize $(+0.31)$}} & 57.64 & \colorbox{rqTwoGain!23.68!white}{{\scriptsize $(+0.25)$}} & 42.73 & \colorbox{rqTwoLoss!45.00!white}{{\scriptsize $(-0.33)$}} & 36.76 & \colorbox{rqTwoGain!40.38!white}{{\scriptsize $(+0.70)$}} \\
\midrule
\quad + \textbf{Item-space OPD (ours)} & \textbf{64.34} & \colorbox{rqTwoGain!45.00!white}{{\scriptsize $(+0.90)$}} & \textbf{57.98} & \colorbox{rqTwoGain!45.00!white}{{\scriptsize $(+0.59)$}} & \textbf{43.55} & \colorbox{rqTwoGain!45.00!white}{{\scriptsize $(+0.49)$}} & \textbf{36.86} & \colorbox{rqTwoGain!45.00!white}{{\scriptsize $(+0.80)$}} \\
\bottomrule
\end{tabular}
\endgroup

\end{adjustbox}
\end{table}

\paragraph{Results.}
As shown in Table~\ref{tab:rq2_objectives}, \keyresult{our method is the only method that consistently improves RankNet across all four benchmarks.} \textbf{Offline KL} improves BRIGHT but degrades general retrieval and R2MED, suggesting that the teacher's first-token distribution does not adequately capture ranking preferences beyond the initial selection. \textbf{AR SFT and AR OPD} improve reasoning-intensive retrieval but hurt general retrieval, consistent with a mismatch between sequence-generation training and first-token-score inference. \textbf{Listwise GRPO} provides ranking-level scalar rewards without explicit feedback on individual selections. In contrast, \textbf{our method} directly supervises the scores used at inference with conditional teacher feedback at each supervised selection step, achieving consistent gains across both general and reasoning-intensive retrieval.

\subsection{RQ3: Does combining offline and on-policy supervision improve ranking?}
\label{sec:rq3}
\paragraph{Component ablation.}
To answer this question, we compare RankNet alone, Item-space OPD alone, and their joint objective (Table~\ref{tab:rq3_ablation}). Zero-shot first-token scoring and sliding-window generation establish the student's performance before distillation.

\begin{table}[!htbp]
\centering
\small
\setlength{\tabcolsep}{4pt}
\renewcommand{\arraystretch}{1.15}
\caption{\textbf{Ablation of offline and on-policy supervision.} All student configurations use Qwen3-8B, with Qwen3.8-27B providing training supervision. The teacher's sliding-window generation is shown separately as a reference. Scores are nDCG@10 $\times 100$, averaged across datasets within each benchmark. Bold and underlined values mark the best and second-best student scores.}
\label{tab:rq3_ablation}

\definecolor{rqThreeOurs}{HTML}{EEE8F7}
\definecolor{rqThreeTeacher}{HTML}{666666}
\begin{tabularx}{\linewidth}{l*{4}{>{\centering\arraybackslash}X}}
\toprule
& \multicolumn{2}{c}{General Retrieval} & \multicolumn{2}{c}{Reasoning-Intensive Retrieval} \\
\cmidrule(lr){2-3}\cmidrule(lr){4-5}
\multicolumn{1}{c}{Configuration} & TREC DL & BEIR & R2MED & BRIGHT \\
\midrule
\textcolor{rqThreeTeacher}{Qwen3.8-27B (teacher)} & \textcolor{rqThreeTeacher}{62.97} & \textcolor{rqThreeTeacher}{57.26} & \textcolor{rqThreeTeacher}{46.66} & \textcolor{rqThreeTeacher}{37.42} \\
\midrule
Zero-shot (first-token) & 47.23 & 46.99 & 31.15 & 30.88 \\
Zero-shot (sliding-window) & 58.31 & 53.40 & 35.96 & 28.42 \\
\midrule
RankNet only & 63.44 & 57.39 & \underline{43.06} & \underline{36.06} \\
Item-space OPD only & \underline{63.80} & \underline{57.93} & 42.75 & 35.85 \\
\rowcolor{rqThreeOurs}
RankNet + Item-space OPD & \textbf{64.34} & \textbf{57.98} & \textbf{43.55} & \textbf{36.86} \\
\bottomrule
\end{tabularx}

\end{table}

\paragraph{Results.}
\keyresult{Joint training achieves the highest score among the student configurations on all four benchmarks.} Adding Item-space OPD to RankNet improves both general and reasoning-intensive retrieval. The joint configuration also outperforms Item-space OPD alone across all four benchmarks, supporting the use of offline ranking supervision together with feedback on the student's current predictions. These results show that the two forms of supervision are complementary: offline rankings provide a consistent training target, and on-policy feedback refines the first-token scores as the student's ranking evolves.

\paragraph{Stability across training seeds.}
\keyresult{The gains from on-policy feedback persist across training seeds.} Table~\ref{tab:rq3_seed_stability} compares RankNet and joint training for the 4B and 8B students on TREC DL and BRIGHT. Joint training improves over RankNet in every matched-seed comparison, covering both student sizes and both benchmarks. Appendix~\ref{app:seed_stability} reports individual runs and paired significance tests.

\begin{table}[!htbp]
\centering
\small
\setlength{\tabcolsep}{5pt}
\renewcommand{\arraystretch}{1.12}
\caption{\textbf{Stability across training seeds.} Qwen3-4B and Qwen3-8B students use the Qwen3.8-27B teacher. We report mean $\pm$ standard deviation of benchmark-average nDCG@10 $\times 100$ across training seeds 42, 43, and 44.}
\label{tab:rq3_seed_stability}

\definecolor{seedStabilityOurs}{HTML}{EEE8F7}
\begin{tabularx}{\linewidth}{ll*{2}{>{\centering\arraybackslash}X}}
\toprule
Student & Objective & TREC DL & BRIGHT \\
\midrule
Qwen3-4B & RankNet & $63.16 \pm 0.02$ & $35.16 \pm 0.34$ \\
\rowcolor{seedStabilityOurs}
 & + Item-space OPD & $\mathbf{64.21} \pm 0.13$ & $\mathbf{35.63} \pm 0.17$ \\
\midrule
Qwen3-8B & RankNet & $63.40 \pm 0.06$ & $36.08 \pm 0.11$ \\
\rowcolor{seedStabilityOurs}
 & + Item-space OPD & $\mathbf{64.21} \pm 0.14$ & $\mathbf{36.85} \pm 0.08$ \\
\bottomrule
\end{tabularx}

\end{table}

\subsection{RQ4: Does our method generalize across teachers and student sizes?}
\label{sec:rq4}
Our item-space alignment design enables students to learn from teachers across model families. To answer this question, we distill from Gemma-4-31B, Qwen3.8-27B, and ReasonRank-32B into Qwen3 students at 1.7B, 4B, and 8B parameters. ReasonRank-32B is built on Qwen2.5-32B-Instruct \citep{liu2026reasonrank}. Table~\ref{tab:rq4_teacher_transfer} compares each student's zero-shot performance with joint RankNet and Item-space OPD training under each teacher.

\begin{table}[!ht]
\centering
\small
\setlength{\tabcolsep}{3pt}
\renewcommand{\arraystretch}{1.20}
\caption{\textbf{Distillation across teacher families and student sizes.} Scores are benchmark-average nDCG@10 $\times 100$. Distilled students use RankNet and Item-space OPD; $\uparrow$ indicates gains over the same row's zero-shot student. Gray teacher scores are generative references. Bold marks the highest student score in each row.}
\label{tab:rq4_teacher_transfer}
\begin{adjustbox}{max width=\linewidth}

\definecolor{rqFourTeacher}{HTML}{737373}
\definecolor{rqFourGain}{HTML}{277A47}
\definecolor{rqFourLoss}{HTML}{B76732}
\begin{tabular}{@{}llc*{3}{cc}@{}}
\toprule
\multirow{2}{*}{Benchmark} & \multirow{2}{*}{Student} & \multirow{2}{*}{Zero-shot} & \multicolumn{2}{c}{Gemma-4-31B} & \multicolumn{2}{c}{Qwen3.8-27B} & \multicolumn{2}{c}{ReasonRank-32B} \\
\cmidrule(lr){4-5}\cmidrule(lr){6-7}\cmidrule(lr){8-9}
& & & \textcolor{rqFourTeacher}{Teacher} & Distilled & \textcolor{rqFourTeacher}{Teacher} & Distilled & \textcolor{rqFourTeacher}{Teacher} & Distilled \\
\midrule
\multirow{3}{*}{TREC DL} & 1.7B & 39.67 & \multirow{3}{*}{\textcolor{rqFourTeacher}{64.38}} & \textbf{64.69} {\scriptsize\textcolor{rqFourGain}{$\uparrow\,25.02$}} & \multirow{3}{*}{\textcolor{rqFourTeacher}{62.97}} & 63.89 {\scriptsize\textcolor{rqFourGain}{$\uparrow\,24.22$}} & \multirow{3}{*}{\textcolor{rqFourTeacher}{62.87}} & 64.20 {\scriptsize\textcolor{rqFourGain}{$\uparrow\,24.53$}} \\
 & 4B & 46.21 &  & \textbf{64.70} {\scriptsize\textcolor{rqFourGain}{$\uparrow\,18.49$}} &  & 64.25 {\scriptsize\textcolor{rqFourGain}{$\uparrow\,18.04$}} &  & 64.27 {\scriptsize\textcolor{rqFourGain}{$\uparrow\,18.06$}} \\
 & 8B & 47.23 &  & \textbf{65.04} {\scriptsize\textcolor{rqFourGain}{$\uparrow\,17.81$}} &  & 64.34 {\scriptsize\textcolor{rqFourGain}{$\uparrow\,17.11$}} &  & 64.86 {\scriptsize\textcolor{rqFourGain}{$\uparrow\,17.63$}} \\
\midrule
\multirow{3}{*}{BEIR} & 1.7B & 33.40 & \multirow{3}{*}{\textcolor{rqFourTeacher}{57.25}} & 55.73 {\scriptsize\textcolor{rqFourGain}{$\uparrow\,22.33$}} & \multirow{3}{*}{\textcolor{rqFourTeacher}{57.26}} & \textbf{56.65} {\scriptsize\textcolor{rqFourGain}{$\uparrow\,23.25$}} & \multirow{3}{*}{\textcolor{rqFourTeacher}{56.23}} & 56.22 {\scriptsize\textcolor{rqFourGain}{$\uparrow\,22.82$}} \\
 & 4B & 43.03 &  & 57.15 {\scriptsize\textcolor{rqFourGain}{$\uparrow\,14.12$}} &  & \textbf{57.60} {\scriptsize\textcolor{rqFourGain}{$\uparrow\,14.57$}} &  & 56.70 {\scriptsize\textcolor{rqFourGain}{$\uparrow\,13.67$}} \\
 & 8B & 46.99 &  & 57.45 {\scriptsize\textcolor{rqFourGain}{$\uparrow\,10.46$}} &  & \textbf{57.98} {\scriptsize\textcolor{rqFourGain}{$\uparrow\,10.99$}} &  & 56.86 {\scriptsize\textcolor{rqFourGain}{$\uparrow\,9.87$}} \\
\midrule
\multirow{3}{*}{R2MED} & 1.7B & 21.16 & \multirow{3}{*}{\textcolor{rqFourTeacher}{47.74}} & 38.89 {\scriptsize\textcolor{rqFourGain}{$\uparrow\,17.73$}} & \multirow{3}{*}{\textcolor{rqFourTeacher}{46.66}} & 38.45 {\scriptsize\textcolor{rqFourGain}{$\uparrow\,17.29$}} & \multirow{3}{*}{\textcolor{rqFourTeacher}{44.21}} & \textbf{39.44} {\scriptsize\textcolor{rqFourGain}{$\uparrow\,18.28$}} \\
 & 4B & 27.60 &  & 43.01 {\scriptsize\textcolor{rqFourGain}{$\uparrow\,15.41$}} &  & 42.39 {\scriptsize\textcolor{rqFourGain}{$\uparrow\,14.79$}} &  & \textbf{43.16} {\scriptsize\textcolor{rqFourGain}{$\uparrow\,15.56$}} \\
 & 8B & 31.15 &  & \textbf{43.92} {\scriptsize\textcolor{rqFourGain}{$\uparrow\,12.77$}} &  & 43.55 {\scriptsize\textcolor{rqFourGain}{$\uparrow\,12.40$}} &  & 43.65 {\scriptsize\textcolor{rqFourGain}{$\uparrow\,12.50$}} \\
\midrule
\multirow{3}{*}{BRIGHT} & 1.7B & 26.68 & \multirow{3}{*}{\textcolor{rqFourTeacher}{39.94}} & 33.66 {\scriptsize\textcolor{rqFourGain}{$\uparrow\,6.98$}} & \multirow{3}{*}{\textcolor{rqFourTeacher}{37.42}} & \textbf{34.08} {\scriptsize\textcolor{rqFourGain}{$\uparrow\,7.40$}} & \multirow{3}{*}{\textcolor{rqFourTeacher}{34.78}} & 33.37 {\scriptsize\textcolor{rqFourGain}{$\uparrow\,6.69$}} \\
 & 4B & 28.81 &  & \textbf{35.70} {\scriptsize\textcolor{rqFourGain}{$\uparrow\,6.89$}} &  & 35.61 {\scriptsize\textcolor{rqFourGain}{$\uparrow\,6.80$}} &  & \textbf{35.70} {\scriptsize\textcolor{rqFourGain}{$\uparrow\,6.89$}} \\
 & 8B & 30.88 &  & \textbf{37.28} {\scriptsize\textcolor{rqFourGain}{$\uparrow\,6.40$}} &  & 36.86 {\scriptsize\textcolor{rqFourGain}{$\uparrow\,5.98$}} &  & 36.05 {\scriptsize\textcolor{rqFourGain}{$\uparrow\,5.17$}} \\
\bottomrule
\end{tabular}

\end{adjustbox}
\end{table}

\paragraph{Results.}
\keyresult{Our method generalizes across teacher families and student sizes, with consistent gains across general and reasoning-intensive retrieval benchmarks.} Each of the three teachers improves all three student sizes over their zero-shot counterparts on all four benchmarks.

\section{Related Work}
\label{sec:related_work}

\subsection{Efficient LLM-based Reranking}

Generative listwise rerankers, such as RankGPT \citep{sun2023chatgpt} and RankZephyr \citep{pradeep2023rankzephyr}, produce rankings autoregressively, incurring decoding latency and repeated model calls for long candidate lists. TourRank \citep{chen2025tourrank} and GroupRank \citep{long2026grouprank} improve efficiency through grouped comparisons and parallel processing. In contrast, FIRST \citep{gangireddy2024first} eliminates ranking-sequence generation by sorting first-token document-ID logits, with subsequent work extending this approach \citep{chen2025accelerating}. Our work builds on first-token ranking and focuses on learning these scores from a generative teacher.

\subsection{Knowledge Distillation for Reranking}

Knowledge distillation transfers ranking knowledge from powerful teachers to smaller models. RankZephyr \citep{pradeep2023rankzephyr} and LiT5-Distill \citep{tamber2023scaling} train generative rerankers on teacher-generated rankings, while Rank-DistiLLM \citep{schlatt2025rankdistillm} distills teacher rankings into cross-encoders using learning-to-rank objectives. DeAR \citep{abdallah2025dear} further incorporates reasoning-based supervision. Beyond offline distillation, GKD \citep{agarwal2024onpolicy} and MiniLLM \citep{gu2024minillm} leverage teacher feedback on student-generated sequences. Unlike these autoregressive approaches, \dr applies on-policy teacher feedback directly to first-token ranking scores.
\section{Conclusion}
We introduced \dr, a decision reranking model that combines offline ranking supervision with item-space on-policy distillation. Conditional teacher feedback on the student's current ranking directly trains the first-token scores used at inference. This approach improves ranking across general and reasoning-intensive retrieval and generalizes across teacher families and student sizes. On TREC DL and BRIGHT, our 8B model outperforms its generative listwise counterpart with substantially lower latency. \dr learns from a generative teacher's sequential ranking decisions while reranking the full candidate set through single-token scoring, offering a promising path to distilling efficient decision models for broader applications.

\bibliographystyle{assets/plainnat}
\bibliography{references}
\clearpage
\beginappendix
\section{Training and Inference Details}
\label{app:training_details}

\paragraph{Training.}
We use a batch size of 32 and a learning rate of $2\times10^{-5}$ for the 1.7B student and $6\times10^{-6}$ for the 4B and 8B students. The joint objective uses the student's top-$K$ candidates with $K=20$, supervises steps $T_{\mathrm{step}}=\{2,\ldots,19\}$, and sets $\lambda=0.1$. Both losses update the student through its first-token scores. Candidate selection follows the student's ranking without positive-document replacement. The OPD-only configuration in Section~\ref{sec:rq3} uses loss weight 1 and supervises steps 1--19.

\paragraph{Teacher supervision.}
All teacher--student combinations use the same training queries and first-token student inference. We obtain offline rankings and teacher reference scores using generative sliding windows of size $W=20$ and stride $s=10$, with thinking disabled for all teachers. The zero-shot student generation reference uses the same window size, stride, and thinking-disabled setting. On-policy feedback uses thinking-disabled configurations for Gemma-4-31B and Qwen3.8-27B, and a thinking-enabled configuration for ReasonRank-32B.

\paragraph{Inference.}
At inference, the student scores the full candidate set in one forward pass. Student inputs truncate each passage to 128 tokens for TREC DL and 384 tokens for BEIR, BRIGHT, and R2MED. The main comparison evaluates baselines using the document budgets and inference parameters of their original implementations; Qwen3-Reranker uses the same passage-length limits as our students.

\paragraph{Jev API evaluation.}
We use Jev (\texttt{jev-1.14.0}) through its Score and Choice interfaces. Score evaluates each query--document pair against 4 ordered relevance grades: 0 (irrelevant), 1 (related but does not answer the query), 2 (highly relevant), and 3 (perfectly relevant). Let $p_i(r)$ denote the probability of grade $r$ for document $d_i$. We rank documents by the API's returned expected relevance score,
\begin{equation}
    s_i^{\mathrm{Jev}}=\sum_{r=0}^{3}r\,p_i(r).
    \label{eq:jev_expected_score}
\end{equation}
Score questions are batched using the same passage-length limits as our students. Choice presents all 100 candidates as options in one question and ranks them by their returned probabilities. Passages are shortened uniformly when needed to fit the API context. Both modes break ties using retrieval order. To make the latency comparison fairer, we use server-reported time to exclude client-side network and connection delays. Queries are processed sequentially, and we report median per-query latency from the \texttt{x-envoy-upstream-service-time} response header, converted from milliseconds to seconds.

\paragraph{Latency measurement.}
For the local models in Section~\ref{sec:rq1_efficiency}, we use each method's original inference configuration with passages truncated to 128 tokens for TREC DL and 384 tokens for BRIGHT, measuring both effectiveness and latency under these settings. We rerank 100 candidates from BM25 and ReasonIR-8B, respectively, using one vLLM worker per NVIDIA B200. After warm-up, queries are processed individually; we take the median latency per worker, then the median across 8 workers. Pointwise rerankers process all 100 query--document pairs concurrently; timing ends when all scores are available. Listwise timings include all required model calls and decoding steps. Student timings cover the forward pass and exclude score sorting.

\Needspace{30\baselineskip}
\subsection{Prompt Examples}
\label{app:prompt_examples}
The following excerpts abbreviate the recorded prompts and requests, using placeholders for query and passage text and simplified chat formatting.

\paragraph{Listwise ranking prompt.}
Both student and teacher inputs contain the query, candidate passages with document IDs, and ranking instructions. The excerpt below shows the student's system instruction, user message, and answer opening. The student reads the first document-ID logits after the opening bracket as ranking scores. Teachers use numerical IDs to generate offline rankings or evaluate the student's selection sequence through teacher forcing.
\begin{Verbatim}[fontsize=\footnotesize,samepage=true,breaklines=true,breakanywhere=true,breaksymbolleft={},frame=single,framesep=4pt]
System:
You are RankLLM, an intelligent assistant that can rank passages based on their relevancy to the query

User:
I will provide you with 100 passages, each indicated by an alphabetical identifier []. Rank the passages based on their relevance to the search query: {query}.
[AA] {passage 1}
[AB] {passage 2}
...
[ET] {passage 100}
Search Query: {query}
Rank the 100 passages above based on their relevance to the search query.
... The output format should be [] > [], e.g., [AB] > [AA], ...

Assistant prefix: <think> </think> [
\end{Verbatim}

\paragraph{Jev request fields.}
The query is shared through \texttt{state}. Score uses one question per passage, while Choice uses one question containing all candidates. The Score criteria are abbreviated to the grade names defined above.
\begin{Verbatim}[fontsize=\footnotesize,samepage=true,breaklines=true,breakanywhere=true,breaksymbolleft={},frame=single,framesep=4pt]
model: "jev-1.14.0"
state: "Query: {query}"

Score:
  type: "score"
  instructions: "How relevant is the following passage to the query?\n\nPassage: {passage}"
  criteria: [Irrelevant, Related, Highly relevant, Perfectly relevant]

Choice:
  type: "choice"
  instructions: "Which passage is the most relevant to the query?"
  criteria: {"1": "{passage 1}", ..., "100": "{passage 100}"}
\end{Verbatim}

\clearpage
\section{Detailed Results under Shared Passage-Length Limits}
\label{app:rq1_detailed}
Tables~\ref{tab:rq1_trec_dl}--\ref{tab:rq1_r2med} report a controlled comparison with shared passage-length limits: 128 tokens for TREC DL and 384 tokens for BEIR, BRIGHT, and R2MED. They include model backbones and references, as well as the additional Qwen3-Reranker and TFRank configurations below 4B parameters. Within the pointwise and listwise groups, methods are ordered by the first public release of their papers. RankMistral denotes the RankMistral100 checkpoint. Bold and underlined values mark the best and second-best scores in each column, respectively.

\begin{table}[!htbp]
\centering
\small
\setlength{\tabcolsep}{4pt}
\renewcommand{\arraystretch}{1.05}
\caption{\textbf{Comparison with existing rerankers on TREC DL.} All rerankers use BM25 top-100 candidates with passages truncated to 128 tokens. Our Qwen3-based \dr students are distilled from Qwen3.8-27B. We report nDCG@10 $\times 100$; Avg. is the mean across the five years.}
\label{tab:rq1_trec_dl}
\begin{adjustbox}{max width=\linewidth}

\begin{tabular}{@{}ll*{6}{c}@{}}
\toprule
& & \multicolumn{2}{c}{MS MARCO v1} & \multicolumn{3}{c}{MS MARCO v2} & \\
\cmidrule(lr){3-4}\cmidrule(lr){5-7}
\multicolumn{1}{c}{Method} & \multicolumn{1}{c}{Backbone} & DL19 & DL20 & DL21 & DL22 & DL23 & Avg. \\
\midrule
\shortstack[l]{BM25\\[-1pt]{\scriptsize\citep{robertson2009probabilistic}}} & -- & 50.58 & 47.96 & 44.58 & 26.92 & 26.24 & 39.26 \\
\midrule
\multicolumn{8}{l}{\textit{Pointwise}} \\
\multirow{3}{*}{\shortstack[l]{Qwen3-Reranker\\[-1pt]{\scriptsize\citep{zhang2025qwen3embedding}}}} & Qwen3-0.6B & 69.42 & 63.55 & 65.48 & 46.00 & 38.96 & 56.68 \\
 & Qwen3-4B & 73.53 & 69.04 & 67.94 & 50.51 & 45.69 & 61.34 \\
 & Qwen3-8B & 73.12 & 65.98 & 66.29 & 51.21 & 45.52 & 60.42 \\
\multirow{4}{*}{\shortstack[l]{TFRank\\[-1pt]{\scriptsize\citep{fan2025tfrank}}}} & Qwen3-0.6B & 66.61 & 64.49 & 57.60 & 40.83 & 36.88 & 53.28 \\
 & Qwen3-1.7B & 68.30 & 62.87 & 59.67 & 41.75 & 37.40 & 54.00 \\
 & Qwen3-4B & 71.81 & 67.56 & 61.23 & 46.74 & 40.40 & 57.55 \\
 & Qwen3-8B & 71.29 & 67.12 & 63.18 & 46.28 & 41.29 & 57.83 \\
\multirow{3}{*}{\shortstack[l]{ERank\\[-1pt]{\scriptsize\citep{cai2025erank}}}} & Qwen3-4B & 72.84 & 67.61 & 67.75 & 45.15 & 43.48 & 59.36 \\
 & Qwen3-14B & 70.11 & 67.40 & 67.00 & 45.08 & 42.86 & 58.49 \\
 & Qwen3-32B & 72.15 & 67.56 & 65.94 & 46.59 & 43.92 & 59.23 \\
\shortstack[l]{DirectRank\\[-1pt]{\scriptsize\citep{lu2026overthinking}}} & Qwen3-8B & 68.96 & 67.86 & 65.48 & 48.66 & 43.03 & 58.80 \\
\midrule
\multicolumn{8}{l}{\textit{Listwise}} \\
\shortstack[l]{RankZephyr\\[-1pt]{\scriptsize\citep{pradeep2023rankzephyr}}} & Zephyr-7B & 73.93 & 69.87 & 70.03 & 51.21 & 45.00 & 62.01 \\
\shortstack[l]{FIRST\\[-1pt]{\scriptsize\citep{gangireddy2024first}}} & Zephyr-7B & 71.84 & 70.79 & 69.14 & 49.77 & 45.13 & 61.33 \\
\shortstack[l]{FirstMistral\\[-1pt]{\scriptsize\citep{chen2025accelerating}}} & Mistral-7B & 72.19 & 70.54 & 67.90 & 49.65 & 45.00 & 61.06 \\
\shortstack[l]{RankMistral\\[-1pt]{\scriptsize\citep{liu2024sliding}}} & Mistral-7B & 72.68 & 71.42 & 69.90 & 48.65 & 45.90 & 61.71 \\
\multirow{2}{*}{\shortstack[l]{Rank-R1\\[-1pt]{\scriptsize\citep{zhuang2025rankr1}}}} & Qwen2.5-7B & 71.78 & 67.67 & 68.68 & 48.26 & 42.24 & 59.73 \\
 & Qwen2.5-14B & 72.54 & 68.91 & 67.86 & 47.60 & 44.66 & 60.31 \\
DirectRank & Qwen3-8B & 74.51 & 70.34 & 70.60 & \textbf{52.33} & 45.92 & 62.74 \\
\multirow{2}{*}{\shortstack[l]{GroupRank\\[-1pt]{\scriptsize\citep{long2026grouprank}}}} & Qwen2.5-7B & 67.19 & 63.28 & 61.42 & 44.02 & 39.51 & 55.08 \\
 & Qwen2.5-32B & 68.81 & 64.46 & 61.18 & 43.27 & 40.81 & 55.71 \\
\midrule
\multirow{3}{*}{\textbf{\dr}} & Qwen3-1.7B & 75.03 & \textbf{72.87} & 71.20 & 51.07 & 49.26 & 63.89 \\
 & Qwen3-4B & \underline{75.48} & 72.38 & \textbf{72.36} & 51.46 & \underline{49.56} & \underline{64.25} \\
 & Qwen3-8B & \textbf{75.63} & \underline{72.61} & \underline{71.45} & \underline{52.27} & \textbf{49.73} & \textbf{64.34} \\
\bottomrule
\end{tabular}

\end{adjustbox}
\end{table}

\begin{table}[!htbp]
\centering
\small
\setlength{\tabcolsep}{3pt}
\renewcommand{\arraystretch}{1.05}
\caption{\textbf{Comparison with existing rerankers on BEIR.} All rerankers use BM25 top-100 candidates with passages truncated to 384 tokens. Our Qwen3-based \dr students are distilled from Qwen3.8-27B. We report nDCG@10 $\times 100$; Avg. is the mean across the displayed datasets.}
\label{tab:rq1_beir}
\begin{adjustbox}{max width=\linewidth}

\begin{tabular}{@{}ll*{8}{c}@{}}
\toprule
\multicolumn{1}{c}{Method} & \multicolumn{1}{c}{Backbone} & COVID & NFCorpus & DBPedia & SciFact & Signal & News & Robust04 & Avg. \\
\midrule
\shortstack[l]{BM25\\[-1pt]{\scriptsize\citep{robertson2009probabilistic}}} & -- & 59.47 & 33.75 & 31.80 & 67.89 & 33.08 & 39.52 & 40.70 & 43.74 \\
\midrule
\multicolumn{10}{l}{\textit{Pointwise}} \\
\multirow{3}{*}{\shortstack[l]{Qwen3-Reranker\\[-1pt]{\scriptsize\citep{zhang2025qwen3embedding}}}} & Qwen3-0.6B & 85.96 & 38.49 & 43.30 & 76.95 & 30.64 & 46.10 & 55.23 & 53.81 \\
 & Qwen3-4B & 84.66 & 39.62 & 46.77 & \underline{79.48} & 33.15 & 48.51 & 57.70 & 55.70 \\
 & Qwen3-8B & 83.68 & 39.58 & 46.24 & 79.41 & 32.59 & 50.74 & 56.64 & 55.55 \\
\multirow{4}{*}{\shortstack[l]{TFRank\\[-1pt]{\scriptsize\citep{fan2025tfrank}}}} & Qwen3-0.6B & 81.04 & 37.35 & 40.83 & 73.70 & 28.72 & 41.71 & 53.64 & 51.00 \\
 & Qwen3-1.7B & 77.48 & 35.99 & 38.66 & 71.91 & 27.22 & 44.26 & 52.25 & 49.68 \\
 & Qwen3-4B & 84.28 & 38.50 & 39.74 & 77.77 & 31.13 & 47.54 & 58.98 & 53.99 \\
 & Qwen3-8B & 84.46 & 38.23 & 42.85 & 76.89 & 29.97 & 48.36 & 59.07 & 54.26 \\
\multirow{3}{*}{\shortstack[l]{ERank\\[-1pt]{\scriptsize\citep{cai2025erank}}}} & Qwen3-4B & 83.29 & 39.12 & 41.20 & 73.80 & 32.22 & 47.59 & 61.30 & 54.07 \\
 & Qwen3-14B & 83.99 & 38.58 & 42.12 & 75.07 & 32.11 & 47.17 & 61.56 & 54.37 \\
 & Qwen3-32B & 82.08 & 38.97 & 41.72 & 75.57 & 30.91 & 46.64 & 61.05 & 53.85 \\
\shortstack[l]{DirectRank\\[-1pt]{\scriptsize\citep{lu2026overthinking}}} & Qwen3-8B & 82.24 & 38.34 & 41.94 & 74.72 & 26.30 & 46.92 & 59.36 & 52.83 \\
\midrule
\multicolumn{10}{l}{\textit{Listwise}} \\
\shortstack[l]{RankZephyr\\[-1pt]{\scriptsize\citep{pradeep2023rankzephyr}}} & Zephyr-7B & 58.54 & 30.65 & 44.32 & 53.01 & 30.47 & 33.42 & 34.95 & 40.77 \\
\shortstack[l]{FIRST\\[-1pt]{\scriptsize\citep{gangireddy2024first}}} & Zephyr-7B & 82.06 & 31.59 & 46.34 & 41.70 & 34.22 & 40.97 & 46.62 & 46.21 \\
\shortstack[l]{FirstMistral\\[-1pt]{\scriptsize\citep{chen2025accelerating}}} & Mistral-7B & 86.26 & 38.18 & 45.68 & 78.34 & 32.07 & 49.66 & 59.11 & 55.61 \\
\shortstack[l]{RankMistral\\[-1pt]{\scriptsize\citep{liu2024sliding}}} & Mistral-7B & 82.92 & 38.32 & 43.45 & 75.87 & 33.58 & 49.66 & 57.83 & 54.52 \\
\multirow{2}{*}{\shortstack[l]{Rank-R1\\[-1pt]{\scriptsize\citep{zhuang2025rankr1}}}} & Qwen2.5-7B & 83.28 & 37.61 & 42.96 & 76.83 & 30.66 & 47.67 & 55.18 & 53.46 \\
 & Qwen2.5-14B & 84.46 & 39.40 & 44.59 & 77.89 & 32.98 & 49.57 & 60.23 & 55.59 \\
DirectRank & Qwen3-8B & 83.58 & 39.75 & 46.05 & 78.15 & 32.80 & 51.37 & 61.28 & 56.14 \\
\multirow{2}{*}{\shortstack[l]{GroupRank\\[-1pt]{\scriptsize\citep{long2026grouprank}}}} & Qwen2.5-7B & 80.32 & 36.34 & 37.14 & 73.63 & 23.40 & 41.73 & 55.37 & 49.70 \\
 & Qwen2.5-32B & 78.72 & 37.40 & 37.36 & 74.96 & 27.05 & 42.04 & 58.41 & 50.85 \\
\midrule
\multirow{3}{*}{\textbf{\dr}} & Qwen3-1.7B & 84.05 & 40.01 & 46.33 & 78.01 & \textbf{36.67} & 52.21 & 59.30 & 56.65 \\
 & Qwen3-4B & \underline{86.42} & \underline{40.33} & \underline{46.78} & 79.41 & \underline{35.36} & \underline{52.49} & \underline{62.41} & \underline{57.60} \\
 & Qwen3-8B & \textbf{86.46} & \textbf{40.79} & \textbf{47.15} & \textbf{80.14} & 35.28 & \textbf{52.85} & \textbf{63.22} & \textbf{57.98} \\
\bottomrule
\end{tabular}

\end{adjustbox}
\end{table}

\begin{table}[!htbp]
\centering
\scriptsize
\setlength{\tabcolsep}{2pt}
\renewcommand{\arraystretch}{1.10}
\caption{\textbf{Comparison with existing rerankers on BRIGHT.} All rerankers use ReasonIR-8B top-100 candidates retrieved with GPT-4 reasoning queries, and receive the original query for reranking. Passages are truncated to 384 tokens. Our Qwen3-based \dr students are distilled from Qwen3.8-27B. Scores are nDCG@10 $\times 100$; Avg. is the mean across the twelve tasks.}
\label{tab:rq1_bright}
\begin{adjustbox}{max width=\linewidth}

\begin{tabular}{@{}ll*{13}{c}@{}}
\toprule
\multicolumn{1}{c}{Method} & \multicolumn{1}{c}{Backbone} & Bio. & Earth & Econ. & Psych. & Robot. & Stack & Sustain & Leet. & Pony & AoPS & TheoQ & TheoT & Avg. \\
\midrule
\shortstack[l]{ReasonIR-8B\\[-1pt]{\scriptsize\citep{shao2025reasonir}}} & -- & 43.37 & 43.00 & 32.65 & 39.57 & 20.82 & 30.73 & 27.34 & 31.68 & 19.55 & 7.37 & 33.93 & 36.68 & 30.56 \\
\midrule
\multicolumn{15}{l}{\textit{Pointwise}} \\
\multirow{3}{*}{\shortstack[l]{Qwen3-Reranker\\[-1pt]{\scriptsize\citep{zhang2025qwen3embedding}}}} & Qwen3-0.6B & 21.03 & 21.92 & 17.34 & 21.56 & 12.73 & 14.82 & 14.47 & 31.69 & 4.95 & 8.76 & 34.95 & 30.86 & 19.59 \\
 & Qwen3-4B & 31.91 & 29.79 & 20.55 & 28.52 & 21.11 & 23.70 & 25.00 & \textbf{39.94} & 16.11 & 9.54 & 39.18 & 34.30 & 26.64 \\
 & Qwen3-8B & 32.44 & 29.78 & 24.62 & 32.98 & 24.52 & 23.05 & 25.93 & 31.25 & 22.17 & 8.90 & 35.75 & 35.13 & 27.21 \\
\multirow{4}{*}{\shortstack[l]{TFRank\\[-1pt]{\scriptsize\citep{fan2025tfrank}}}} & Qwen3-0.6B & 25.70 & 24.73 & 20.72 & 25.91 & 15.33 & 11.75 & 19.78 & 24.10 & 28.03 & 2.40 & 23.46 & 23.86 & 20.48 \\
 & Qwen3-1.7B & 35.68 & 34.48 & 22.26 & 28.60 & 17.08 & 20.72 & 22.28 & 17.60 & 31.23 & 3.69 & 27.67 & 31.89 & 24.43 \\
 & Qwen3-4B & 42.07 & 37.48 & 23.61 & 38.02 & 20.84 & 22.34 & 31.94 & 20.06 & 32.87 & 2.24 & 27.15 & 39.15 & 28.15 \\
 & Qwen3-8B & 43.18 & 38.07 & 24.63 & 31.90 & 19.17 & 20.65 & 30.96 & 26.27 & \underline{48.24} & 4.31 & 27.25 & 36.74 & 29.28 \\
\multirow{3}{*}{\shortstack[l]{ERank\\[-1pt]{\scriptsize\citep{cai2025erank}}}} & Qwen3-4B & 42.22 & 39.72 & 29.65 & 40.47 & 23.53 & 27.44 & 33.78 & 35.34 & 25.11 & 10.27 & 37.00 & \textbf{45.46} & 32.50 \\
 & Qwen3-14B & 41.67 & 40.19 & 27.78 & 39.00 & 20.98 & 24.85 & 34.27 & 33.32 & 27.88 & 10.24 & 36.58 & 43.94 & 31.73 \\
 & Qwen3-32B & 46.13 & 40.50 & 29.73 & 37.81 & 23.18 & 26.23 & 38.36 & 35.12 & 26.63 & \textbf{11.79} & 36.78 & 44.43 & 33.06 \\
\shortstack[l]{DirectRank\\[-1pt]{\scriptsize\citep{lu2026overthinking}}} & Qwen3-8B & 41.78 & 42.15 & 26.99 & 36.65 & 21.57 & 21.80 & 32.02 & 27.52 & \textbf{52.28} & 5.85 & 28.72 & 27.48 & 30.40 \\
\midrule
\multicolumn{15}{l}{\textit{Listwise}} \\
\shortstack[l]{RankZephyr\\[-1pt]{\scriptsize\citep{pradeep2023rankzephyr}}} & Zephyr-7B & 40.58 & 28.34 & 21.71 & 27.18 & 11.65 & 20.34 & 25.60 & 30.26 & 23.35 & 6.04 & 26.44 & 22.60 & 23.67 \\
\shortstack[l]{FIRST\\[-1pt]{\scriptsize\citep{gangireddy2024first}}} & Zephyr-7B & 38.76 & 29.96 & 22.18 & 28.53 & 12.05 & 13.34 & 24.55 & 19.36 & 21.45 & 5.52 & 22.21 & 19.59 & 21.46 \\
\shortstack[l]{FirstMistral\\[-1pt]{\scriptsize\citep{chen2025accelerating}}} & Mistral-7B & 44.28 & 36.10 & 23.57 & 35.57 & 14.63 & 16.24 & 28.12 & 29.71 & 34.27 & 9.03 & 30.28 & 30.34 & 27.68 \\
\shortstack[l]{RankMistral\\[-1pt]{\scriptsize\citep{liu2024sliding}}} & Mistral-7B & 45.15 & 35.25 & 25.74 & 38.06 & 18.39 & 22.83 & 32.91 & 29.22 & 19.38 & 9.21 & 33.16 & 38.00 & 28.94 \\
\multirow{2}{*}{\shortstack[l]{Rank-R1\\[-1pt]{\scriptsize\citep{zhuang2025rankr1}}}} & Qwen2.5-7B & 44.06 & 36.01 & 27.94 & 39.40 & 21.38 & 24.68 & 33.79 & 20.93 & 14.33 & 4.37 & 25.81 & 41.52 & 27.85 \\
 & Qwen2.5-14B & 47.68 & 41.56 & 27.94 & 43.70 & 25.26 & 23.72 & 37.79 & 21.95 & 19.07 & 8.20 & 35.20 & 39.46 & 30.96 \\
DirectRank & Qwen3-8B & \textbf{54.44} & 41.65 & \underline{33.23} & 47.52 & \underline{30.19} & 30.00 & 42.98 & 25.42 & 25.18 & 7.31 & 38.08 & 42.13 & 34.84 \\
\multirow{2}{*}{\shortstack[l]{GroupRank\\[-1pt]{\scriptsize\citep{long2026grouprank}}}} & Qwen2.5-7B & 39.02 & 31.22 & 24.50 & 35.38 & 20.35 & 21.79 & 31.61 & 24.02 & 12.07 & 7.02 & 33.07 & 35.35 & 26.28 \\
 & Qwen2.5-32B & 39.14 & 33.87 & 29.08 & 40.07 & 21.36 & 29.83 & 34.73 & 23.62 & 16.19 & 9.30 & 31.18 & 38.57 & 28.91 \\
\midrule
\multirow{3}{*}{\textbf{\dr}} & Qwen3-1.7B & 48.38 & 41.25 & 32.18 & 44.55 & 26.77 & 28.00 & 40.03 & 34.94 & 19.99 & 8.16 & 40.17 & \underline{44.49} & 34.08 \\
 & Qwen3-4B & 49.52 & \underline{43.44} & 32.52 & \underline{47.87} & 28.39 & \underline{31.51} & \textbf{44.97} & \underline{37.36} & 19.97 & 9.66 & \underline{41.12} & 40.98 & \underline{35.61} \\
 & Qwen3-8B & \underline{49.87} & \textbf{45.43} & \textbf{35.02} & \textbf{48.44} & \textbf{31.90} & \textbf{31.59} & \underline{44.80} & 36.02 & 23.14 & \underline{10.50} & \textbf{41.37} & 44.29 & \textbf{36.86} \\
\bottomrule
\end{tabular}

\end{adjustbox}
\end{table}

\FloatBarrier
\begin{table}[!t]
\centering
\small
\setlength{\tabcolsep}{2.5pt}
\renewcommand{\arraystretch}{1.05}
\caption{\textbf{Comparison with existing rerankers on R2MED.} All rerankers use E5-Mistral-7B-Instruct top-100 candidates with passages truncated to 384 tokens. Our Qwen3-based \dr students are distilled from Qwen3.8-27B. Scores are nDCG@10 $\times 100$; Avg. is the mean across the eight datasets.}
\label{tab:rq1_r2med}
\begin{adjustbox}{max width=\linewidth}

\begin{tabular}{@{}ll*{9}{c}@{}}
\toprule
\multicolumn{1}{c}{Method} & \multicolumn{1}{c}{Backbone} & Biology & Bioinfo & MedSci & MedXpert & MedQA & \shortstack{PMC\\Treat} & \shortstack{PMC\\Clin} & IIYi & Avg. \\
\midrule
\shortstack[l]{E5-Mistral\\[-1pt]{\scriptsize\citep{wang2024improving}}} & Mistral-7B & 18.28 & 41.47 & 41.57 & 6.44 & 12.83 & 24.60 & 30.97 & 21.37 & 24.69 \\
\midrule
\multicolumn{11}{l}{\textit{Pointwise}} \\
\multirow{3}{*}{\shortstack[l]{Qwen3-Reranker\\[-1pt]{\scriptsize\citep{zhang2025qwen3embedding}}}} & Qwen3-0.6B & 18.75 & 44.61 & 41.38 & 8.05 & 13.71 & 41.01 & 31.09 & 19.82 & 27.30 \\
 & Qwen3-4B & 28.15 & 55.12 & 49.68 & 14.29 & 26.31 & 42.98 & 36.53 & 24.53 & 34.70 \\
 & Qwen3-8B & 29.97 & 56.80 & 52.89 & 13.90 & 27.35 & 44.57 & 42.40 & 27.98 & 36.98 \\
\multirow{4}{*}{\shortstack[l]{TFRank\\[-1pt]{\scriptsize\citep{fan2025tfrank}}}} & Qwen3-0.6B & 24.03 & 40.93 & 45.53 & 7.66 & 12.69 & 20.73 & 18.93 & 15.51 & 23.25 \\
 & Qwen3-1.7B & 30.62 & 47.07 & 46.49 & 11.38 & 17.50 & 27.05 & 23.90 & 23.28 & 28.41 \\
 & Qwen3-4B & 36.73 & 57.48 & 57.51 & 14.29 & 22.46 & 42.01 & 31.03 & 26.00 & 35.94 \\
 & Qwen3-8B & 38.08 & 55.01 & 54.24 & 11.92 & 18.98 & 42.45 & 30.89 & 23.88 & 34.43 \\
\multirow{3}{*}{\shortstack[l]{ERank\\[-1pt]{\scriptsize\citep{cai2025erank}}}} & Qwen3-4B & 31.95 & 59.48 & 57.33 & 14.92 & 26.54 & 48.23 & 38.62 & 29.08 & 38.27 \\
 & Qwen3-14B & 31.36 & 62.57 & 59.33 & 14.73 & 28.90 & 46.68 & 35.42 & 30.02 & 38.63 \\
 & Qwen3-32B & 34.16 & 64.11 & 62.78 & 14.89 & 26.70 & 47.72 & 35.15 & 27.52 & 39.13 \\
\shortstack[l]{DirectRank\\[-1pt]{\scriptsize\citep{lu2026overthinking}}} & Qwen3-8B & 34.08 & 61.56 & 57.42 & 16.31 & 30.10 & \underline{52.31} & 34.66 & 26.76 & 39.15 \\
\midrule
\multicolumn{11}{l}{\textit{Listwise}} \\
\shortstack[l]{RankZephyr\\[-1pt]{\scriptsize\citep{pradeep2023rankzephyr}}} & Zephyr-7B & 25.22 & 34.20 & 46.49 & 6.71 & 12.73 & 19.53 & 13.42 & 19.55 & 22.23 \\
\shortstack[l]{FIRST\\[-1pt]{\scriptsize\citep{gangireddy2024first}}} & Zephyr-7B & 26.92 & 42.05 & 50.25 & 7.73 & 12.66 & 25.05 & 14.35 & 10.96 & 23.75 \\
\shortstack[l]{FirstMistral\\[-1pt]{\scriptsize\citep{chen2025accelerating}}} & Mistral-7B & 31.79 & 50.04 & 53.06 & 11.92 & 17.01 & 36.94 & 26.96 & 17.30 & 30.63 \\
\shortstack[l]{RankMistral\\[-1pt]{\scriptsize\citep{liu2024sliding}}} & Mistral-7B & 30.33 & 52.14 & 57.04 & 11.27 & 19.00 & 43.54 & 34.10 & 22.20 & 33.70 \\
\multirow{2}{*}{\shortstack[l]{Rank-R1\\[-1pt]{\scriptsize\citep{zhuang2025rankr1}}}} & Qwen2.5-7B & 33.21 & 51.95 & 54.53 & 14.23 & 22.26 & 46.53 & 27.46 & 22.78 & 34.12 \\
 & Qwen2.5-14B & 37.40 & 54.73 & 58.62 & 16.38 & 26.95 & 50.74 & 37.33 & 28.07 & 38.78 \\
DirectRank & Qwen3-8B & \textbf{42.51} & \underline{65.40} & \textbf{68.82} & \textbf{18.61} & \underline{30.23} & 51.93 & 41.38 & \textbf{33.94} & \textbf{44.10} \\
\multirow{2}{*}{\shortstack[l]{GroupRank\\[-1pt]{\scriptsize\citep{long2026grouprank}}}} & Qwen2.5-7B & 32.51 & 48.95 & 52.06 & 12.24 & 17.12 & 41.87 & 21.56 & 16.93 & 30.41 \\
 & Qwen2.5-32B & 36.67 & 53.94 & 54.72 & 14.50 & 25.74 & 44.18 & 31.62 & 25.96 & 35.92 \\
\midrule
\multirow{3}{*}{\textbf{\dr}} & Qwen3-1.7B & 34.94 & 58.93 & 60.74 & 13.91 & 24.19 & 50.25 & 37.97 & 26.68 & 38.45 \\
 & Qwen3-4B & 40.70 & 64.02 & \underline{64.70} & 17.69 & 27.23 & 51.79 & \underline{42.58} & 30.38 & 42.39 \\
 & Qwen3-8B & \underline{41.89} & \textbf{65.63} & 64.35 & \underline{18.09} & \textbf{30.54} & \textbf{53.66} & \textbf{43.04} & \underline{31.19} & \underline{43.55} \\
\bottomrule
\end{tabular}

\end{adjustbox}
\end{table}

\section{Training Stability and Score Calibration}
\label{app:additional_analysis}

\subsection{Results across Training Seeds}
\label{app:seed_stability}
Table~\ref{tab:seed_stability_details} reports individual runs for Qwen3-4B and Qwen3-8B with the Qwen3.8-27B teacher. Training data and evaluation settings are shared across seeds 42, 43, and 44. For paired significance tests, each query's scores are averaged across the 3 seeds before comparing the methods.

\begin{table}[!htbp]
\centering
\small
\setlength{\tabcolsep}{3.5pt}
\renewcommand{\arraystretch}{1.12}
\caption{\textbf{Individual training runs.} Scores are benchmark-average nDCG@10 $\times 100$. Paired $p$-values compare joint training with RankNet using per-query scores averaged across the 3 seeds.}
\label{tab:seed_stability_details}
\begin{adjustbox}{max width=\linewidth}

\definecolor{seedStabilityOurs}{HTML}{EEE8F7}
\begin{tabularx}{\linewidth}{ll*{6}{>{\centering\arraybackslash}X}c}
\toprule
& & \multicolumn{3}{c}{RankNet} & \multicolumn{3}{c}{+ Item-space OPD} & \\
\cmidrule(lr){3-5}\cmidrule(lr){6-8}
Student & Benchmark & 42 & 43 & 44 & 42 & 43 & 44 & $p$ \\
\midrule
Qwen3-4B & TREC DL & 63.18 & 63.13 & 63.17 & \cellcolor{seedStabilityOurs}64.25 & \cellcolor{seedStabilityOurs}64.32 & \cellcolor{seedStabilityOurs}64.06 & $<0.001$ \\
 & BRIGHT & 35.45 & 35.24 & 34.79 & \cellcolor{seedStabilityOurs}35.61 & \cellcolor{seedStabilityOurs}35.81 & \cellcolor{seedStabilityOurs}35.48 & $0.009$ \\
\midrule
Qwen3-8B & TREC DL & 63.44 & 63.41 & 63.33 & \cellcolor{seedStabilityOurs}64.34 & \cellcolor{seedStabilityOurs}64.05 & \cellcolor{seedStabilityOurs}64.24 & $<0.001$ \\
 & BRIGHT & 36.06 & 36.20 & 35.98 & \cellcolor{seedStabilityOurs}36.86 & \cellcolor{seedStabilityOurs}36.77 & \cellcolor{seedStabilityOurs}36.92 & $<0.001$ \\
\bottomrule
\end{tabularx}

\end{adjustbox}
\end{table}

\subsection{Pairwise Score Calibration}
\label{app:calibration}
\paragraph{Pairwise predictions and confidence.}
We evaluate pairs of candidate documents from the same query whose annotated relevance grades differ, treating unjudged documents as grade 0. Let $g_i$ denote the relevance grade of $d_i$, and let $u_{ij}$ be the probability that the model assigns to preferring $d_i$ over $d_j$. Values above $1/2$ favor $d_i$, while values below $1/2$ favor $d_j$. We record whether the predicted preference agrees with the relevance judgments:
\begin{equation}
    y_{ij}=\mathbf{1}\!\left[(u_{ij}>1/2)=(g_i>g_j)\right].
    \label{eq:calibration_pair}
\end{equation}
Its confidence $c_{ij}$ is the probability assigned to the preferred document. We express the strength of this preference through the absolute log-odds $\delta_{ij}$:
\begin{equation}
    \delta_{ij}=\left|\log\frac{u_{ij}}{1-u_{ij}}\right|,
    \qquad c_{ij}=\max(u_{ij},1-u_{ij})=\sigma(\delta_{ij}),
    \label{eq:calibration_confidence}
\end{equation}
where $\sigma(x)=1/(1+\exp(-x))$ is the sigmoid function. A gap of $\delta_{ij}=0$ corresponds to confidence $1/2$; larger gaps indicate stronger preferences. Calibration measures how closely these confidence values match the frequency of correct predictions. We obtain $u_{ij}$ from each model as follows.

\paragraph{\dr students.}
The student assigns each document a first-token document-ID logit $s_i$. We convert the difference between two logits into a pairwise preference probability, as in the RankNet objective:
\begin{equation}
    u_{ij}=\sigma(s_i-s_j),
    \qquad \delta_{ij}=|s_i-s_j|,
    \qquad c_{ij}=\sigma(|s_i-s_j|).
    \label{eq:calibration_student}
\end{equation}
Thus, the student's confidence is determined by how far apart its two ranking scores are.

\paragraph{Generative teachers.}
For Qwen3.8-27B and Gemma-4-31B, let $p_i$ denote the probability assigned to document ID $i$ at the first output position after reading all candidates. We normalize the two documents' probabilities to obtain their pairwise preference:
\begin{equation}
    u_{ij}=\frac{p_i}{p_i+p_j}=\sigma(\log p_i-\log p_j),
    \qquad c_{ij}=\frac{\max(p_i,p_j)}{p_i+p_j}.
    \label{eq:calibration_teacher}
\end{equation}
Here, $\delta_{ij}=|\log p_i-\log p_j|$. These probabilities measure the teacher's preference between the two documents at its initial selection.

\paragraph{Pointwise rerankers.}
DirectRank and ERank score each query--document pair separately. Let $z_i$ denote the relevance score for document $d_i$. For DirectRank, this score is the log-odds of the relevance labels \texttt{true} and \texttt{false}:
\begin{equation}
    z_i=\log\frac{P(\mathrm{true}\mid q,d_i)}{P(\mathrm{false}\mid q,d_i)}.
    \label{eq:calibration_pointwise_score}
\end{equation}
For ERank, $z_i$ is its predicted document relevance score. Both models use the same conversion from score differences to pairwise probabilities and confidence:
\begin{equation}
    u_{ij}=\sigma(z_i-z_j),
    \qquad \delta_{ij}=|z_i-z_j|,
    \qquad c_{ij}=\sigma(|z_i-z_j|).
    \label{eq:calibration_pointwise}
\end{equation}
A larger score difference produces a more confident pairwise preference. Both pointwise baselines in Figure~\ref{fig:rq1_calibration} use this readout and the same binning and ECE calculation described below.

\paragraph{Jev.}
Jev's Score interface returns a distribution $p_i(a)$ over 4 relevance grades, $a\in\{0,1,2,3\}$, for each document (Appendix~\ref{app:training_details}). Treating the predicted grades as independent, we form their joint distribution as $p_i(a)p_j(b)$ and compute
\begin{equation}
    u_{ij}=\sum_{a>b}p_i(a)p_j(b)+\frac12\sum_a p_i(a)p_j(a).
    \label{eq:calibration_jev}
\end{equation}
The first term sums the probability of $d_i$ receiving a higher grade than $d_j$. The second splits the probability of equal predicted grades evenly between the two preferences. We then compute $\delta_{ij}$ and $c_{ij}=\max(u_{ij},1-u_{ij})$ using Equation~\ref{eq:calibration_confidence}. Jev's pairwise confidence therefore uses its full grade distributions. Its ranking scores use the expected grade in Equation~\ref{eq:jev_expected_score}.

\paragraph{Binning and ECE.}
We examine calibration as a function of the score gap, following \citet{peng2023stabilizing}. We group pairs by $\delta_{ij}$ into bins $[0,0.5),[0.5,1),\ldots,[7.5,8)$ and $[8,\infty)$. For each nonempty bin $B_b$, we compute mean confidence and empirical accuracy:
\begin{equation}
    \operatorname{conf}(B_b)=\frac{1}{|B_b|}\sum_{(i,j)\in B_b}c_{ij},
    \qquad
    \operatorname{acc}(B_b)=\frac{1}{|B_b|}\sum_{(i,j)\in B_b}y_{ij}.
    \label{eq:calibration_bin_statistics}
\end{equation}
ECE is the absolute difference between these quantities, averaged over bins with weights proportional to their pair counts (Equation~\ref{eq:calibration_ece}). We pool valid pairs across all queries within each benchmark, so queries with more eligible pairs contribute more weight. Overall pairwise accuracy is the fraction of correct predictions across these pairs. Each model uses the valid pairs available in its recorded outputs; all models use the same evaluation queries and relevance judgments.

\begin{figure}[!htbp]
\centering
\includegraphics[width=\linewidth]{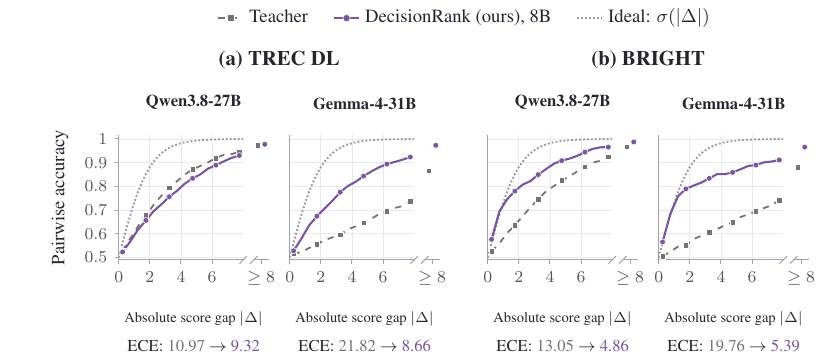}
\caption{\textbf{Calibration of first-token ranking scores on (a) TREC DL and (b) BRIGHT.} Each benchmark compares Qwen3.8-27B and Gemma-4-31B with their respective Qwen3-8B \dr students. The dotted sigmoid curve indicates ideal calibration. Pairwise ECE ($\times100$) is shown as teacher $\rightarrow$ student. These curves use all evaluation queries, matching Figure~\ref{fig:rq1_calibration}.}
\label{fig:calibration_gap}
\end{figure}

\begin{figure}[!htbp]
\centering
\includegraphics[width=\linewidth]{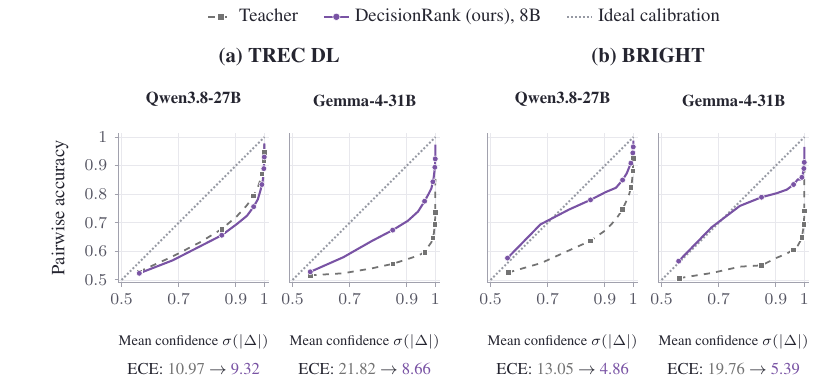}
\caption{\textbf{Reliability on (a) TREC DL and (b) BRIGHT.} Teachers and students match Figure~\ref{fig:calibration_gap}. Points show mean predicted confidence and empirical accuracy within each score-gap bin. The diagonal indicates ideal calibration; ECE ($\times100$) is shown as teacher $\rightarrow$ student.}
\label{fig:calibration_reliability}
\end{figure}

\begin{figure}[!htbp]
\centering
\includegraphics[width=\linewidth]{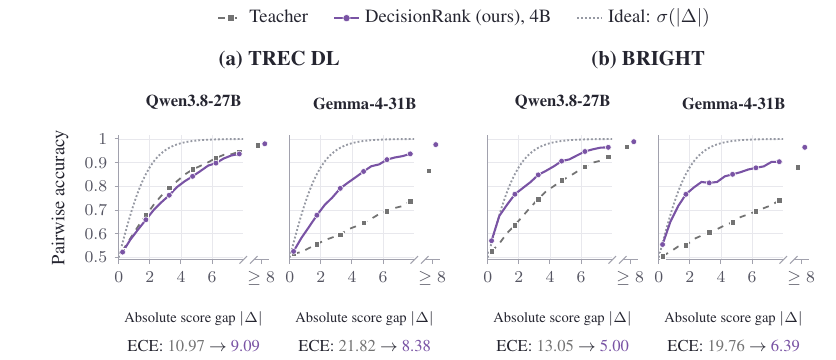}
\caption{\textbf{Calibration of Qwen3-4B students on (a) TREC DL and (b) BRIGHT.} Each teacher is compared with its Qwen3-4B \dr student, using the same settings as Figure~\ref{fig:calibration_gap}. The dotted sigmoid curve indicates ideal calibration; ECE ($\times100$) is shown as teacher $\rightarrow$ student.}
\label{fig:calibration_gap_4b}
\end{figure}

\begin{figure}[!htbp]
\centering
\includegraphics[width=\linewidth]{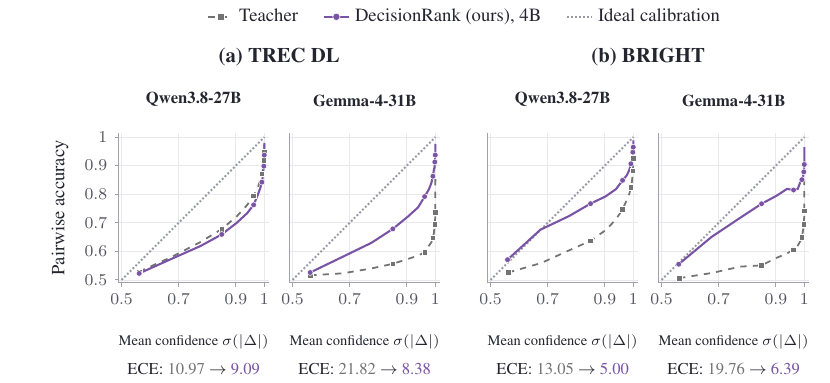}
\caption{\textbf{Reliability of Qwen3-4B students on (a) TREC DL and (b) BRIGHT.} The teachers and students match Figure~\ref{fig:calibration_gap_4b}. Points show mean predicted confidence and empirical accuracy within each score-gap bin. The diagonal indicates ideal calibration; ECE ($\times100$) is shown as teacher $\rightarrow$ student.}
\label{fig:calibration_reliability_4b}
\end{figure}

\paragraph{Full-benchmark score calibration.}
Figures~\ref{fig:calibration_gap}--\ref{fig:calibration_reliability_4b} use all evaluation queries and the original model scores. They compare the Qwen3.8-27B and Gemma-4-31B teachers with their respective 8B and 4B students. The score-gap plots use $|\Delta|=\delta_{ij}$ on the horizontal axis, with $\Delta=s_i-s_j$ for students and $\Delta=\log p_i-\log p_j$ for teachers. Finite score-gap bins are plotted at their midpoints, with the $\delta\geq8$ bin shown separately. Both student sizes have lower ECE and higher pairwise accuracy than their respective teachers on TREC DL and BRIGHT.

\paragraph{Calibration of 4B students.}
The same pattern extends to Qwen3-4B: under both teachers, the student achieves lower aggregate ECE and higher overall pairwise accuracy on TREC DL and BRIGHT. Figures~\ref{fig:calibration_gap_4b} and~\ref{fig:calibration_reliability_4b} use the same binning, axis ranges, and visual layout as the 8B plots, enabling direct comparison across student sizes.

\FloatBarrier

\paragraph{Results.}
\keyresult{The calibration gains are consistent across teacher families and student sizes.} On TREC DL and BRIGHT, the 4B and 8B \dr students trained with either Qwen3.8-27B or Gemma-4-31B achieve lower ECE and higher pairwise accuracy than their corresponding teachers. \dr achieves this improved calibration through ranking supervision alone, without a dedicated calibration objective.

\end{document}